\documentclass[11pt]{article}

\usepackage[final]{acl}

\usepackage{times}
\usepackage{latexsym}

\usepackage[T1]{fontenc}
\usepackage[T5]{fontenc}

\usepackage{subcaption}

\usepackage[utf8]{inputenc}

\usepackage{microtype}

\usepackage{inconsolata}

\usepackage{graphicx}
\usepackage{pifont}

\usepackage{amsmath, amsfonts, amssymb}
\usepackage{multirow}
\usepackage{booktabs}

\usepackage[ruled, lined, linesnumbered]{algorithm2e}
\SetKwComment{Comment}{/* }{ */}
\SetKwProg{Fn}{Function}{:}{end}

\usepackage[tone]{tipa}

\title{Phonetic Modeling of Dialectal Variation in Vietnamese Speech}

\author{
 \textbf{Quan Ngoc Hoang\textsuperscript{1,3,4}},
 \textbf{Long Hoang Huu Nguyen\textsuperscript{1,3,4}}, \\
 \textbf{Nghia Hieu Nguyen\textsuperscript{2,3,4}},
 \textbf{Kiet Van Nguyen\textsuperscript{2,3,4}},
 \textbf{Ngan Luu-Thuy Nguyen\textsuperscript{2,3,4}},
\\
\\
 \textsuperscript{1}Faculty of Computer Science, \\
 \textsuperscript{2}Faculty of Information Science and Engineering, \\
 \textsuperscript{3}University of Information Technology, \\
 \textsuperscript{4}Vietnam National University, Ho Chi Minh city, Vietnam, \\
 \texttt{\{22521178, 22520817\}@gm.uit.edu.vn},  \texttt{\{nghiangh, kietnv,ngannlt\}@uit.edu.vn}\\
}

\begin{document}
\maketitle
\begin{abstract}
Vietnamese exhibits substantial dialectal phonetic variation across Northern, Central, and Southern regions, where identical lexical items may be realized with markedly different pronunciations. Such variation poses challenges for automatic speech recognition (ASR) and remains difficult to model computationally due to the complex relationship between Vietnamese orthography and phonology. Existing approaches typically address dialect variability at the word level, assuming dialect-invariant mappings between spelling and pronunciation, which limits their ability to capture systematic phonetic differences. We propose a dialect-aware phonetic framework that explicitly models Vietnamese phonological structure and dialectal variation at both the vocabulary and decoding levels. The framework introduces a phonetic vocabulary that decomposes each syllable into structured phonetic components and maps them to dialect-specific IPA representations, together with a phonetic-structure decoder that jointly predicts these components. Experiments on the UIT-ViMD, a only-available dataset for multi-dialect in Vietnamese, show that the proposed approach outperforms various pre-trained baselines, \textbf{especially matches the performance of the strongest pretrained wav2ve2-base-vi-250h} across dialects while \textbf{using substantially fewer parameters and no external pretraining}. Code for experimental reproducibility will be publicly available upon the acceptance of this paper.
\end{abstract}

\section{Introduction}

\begin{figure}[t]
    \centering
    \includegraphics[width=\linewidth]{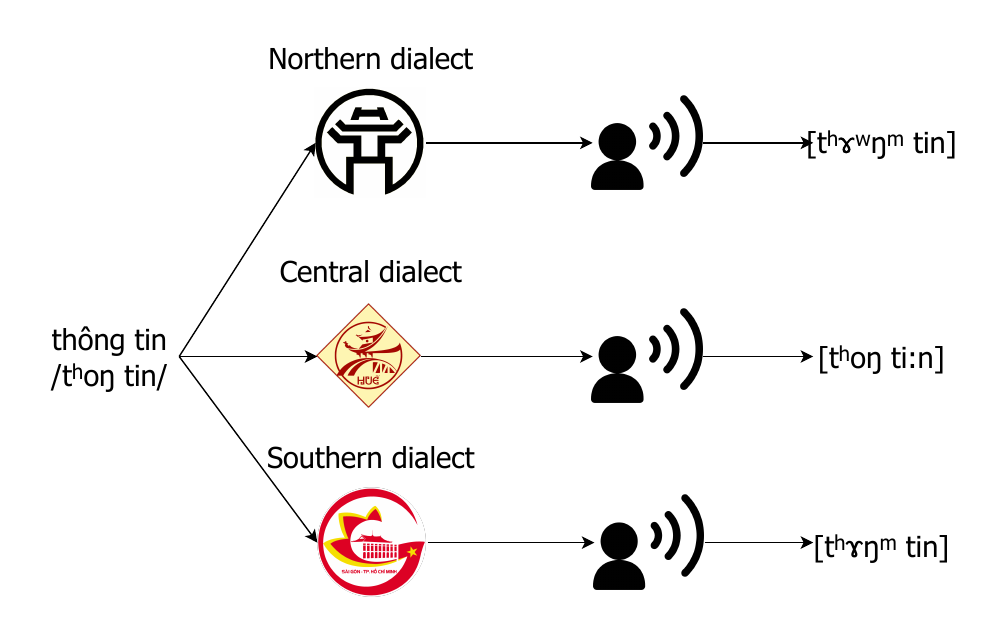}
    \caption{Examples of Vietnamese multi-dialect speech variations.}
    \label{fig:multi-dialect-examples}
\end{figure}

Automatic speech recognition (ASR) systems often experience significant performance degradation when applied to dialectal speech. This challenge is particularly pronounced in Vietnamese, where speakers across Northern, Central, and Southern regions exhibit systematic pronunciation differences despite sharing the same orthographic system (Figure \ref{fig:multi-dialect-examples}). Identical lexical items may therefore correspond to multiple phonetic realizations depending on the dialect, creating a mismatch between written forms and spoken signals.

Most existing ASR systems address dialect variation primarily at the acoustic level. Common strategies include multi-task learning with dialect identification, dialect-specific models, or transfer learning from standard-accent corpora. While these approaches improve robustness, they typically treat dialect differences as acoustic variability to be compensated during training. As a result, the underlying phonological structure and systematic pronunciation shifts across dialects remain largely underexploited.

Vietnamese provides an interesting case for exploring structure-aware modeling. Vietnamese syllables follow a relatively regular phonological template and exhibit consistent relationships between phonetic components and orthography. At the same time, many dialectal differences arise from systematic shifts in specific syllable components, such as initial consonant mergers, vowel quality variation within rhymes, or tone realization. These properties suggest that explicitly modeling phonological structure may provide a more principled way to capture dialect variation.

In this work, we propose a dialect-aware phonetic framework for Vietnamese ASR that incorporates phonological structure directly into the vocabulary representation and decoding process. Instead of predicting orthographic tokens, the model represents each syllable using structured phonetic components and jointly predicts these components during decoding. This design allows the model to capture systematic pronunciation patterns while maintaining interpretable representations that align with Vietnamese phonology.

We implement this idea through a dialect-aware phonetic vocabulary constructed using hierarchical phonetic converters and a phonetic-structure decoder that predicts syllable components in parallel. Experiments on the UIT-ViMD dataset demonstrate that the proposed approach improves recognition performance across dialects while using fewer parameters and no external pretraining. Our main contributions are summarized as follows:

\begin{itemize}
\item We introduce a dialect-aware phonetic vocabulary for Vietnamese that explicitly models syllable structure and systematic dialectal pronunciation shifts.
\item We propose a phonetic-structure decoder that jointly predicts syllable components, enabling structure-aware speech recognition.
\item We demonstrate that phonological modeling improves multi-dialect Vietnamese ASR, achieving superior performance without relying on large-scale pretrained models.
\end{itemize}

Our results suggest that linguistically grounded phonological representations can complement large-scale pretraining, particularly for languages where systematic pronunciation variation occurs at the phonetic level.

\section{Related Work}

Research on dialectal ASR has largely focused on mitigating performance degradation caused by pronunciation variation across regional speech communities. A prominent line of work explores multi-task and multi-model architectures that incorporate dialect labels during training. Multi-task learning frameworks jointly optimize dialect classification and speech recognition objectives to better capture dialect-sensitive acoustic cues~\cite{Multitask_DanZBWJ22}. Similarly, systems equipped with dialect identification modules route utterances to dialect-specific ASR models, reducing cross-dialect interference through specialized decoding paths~\cite{Multitask_ElfekyBVMW16}. While these approaches effectively exploit dialect labels, they typically require large-scale annotated corpora and focus primarily on modeling dialect differences at the acoustic level.

Another line of research addresses dialectal data scarcity and adaptation through transfer learning and multi-stage training strategies. In this setting, models pretrained on high-resource standard-accent corpora are adapted to dialectal speech through fine-tuning or domain adaptation~\cite{Transfer_TaLD24, Transfer_SuwanbanditNSC23}. More recent architectures adopt progressive training schedules to capture diverse acoustic patterns~\cite{Aformer_WangLLW23}. Mixture-of-Experts (MoE) frameworks further improve cross-dialect generalization by dynamically combining dialect-specific and general acoustic experts~\cite{MoE_ZhouGYDW24}. Despite their effectiveness, these approaches largely conceptualize dialectal variation as acoustic variability to be normalized or compensated for, and therefore do not explicitly model the phonological structure underlying dialect-specific pronunciation patterns.

For Vietnamese, recent efforts have focused on constructing multi-dialect speech corpora that enable systematic investigation of regional pronunciation variation~\cite{VietASR_Tran2024, ViMD_DinhDNN24}. Resources such as the UIT-ViMD dataset~\cite{ViMD_DinhDNN24} provide large-scale recordings from speakers across multiple provinces, supporting both speech recognition research and computational analysis of dialectal variation. Nevertheless, most Vietnamese ASR systems continue to operate on orthographic word or subword units~\cite{SurveyVNASR_Nga2021}. These units assume a stable mapping between written forms and pronunciations, an assumption that is often violated in Vietnamese due to systematic cross-dialect phonological shifts involving consonants, vowels, and tones.

Overall, existing work primarily addresses dialectal variation through acoustic modeling and adaptation techniques. In contrast, computational approaches that explicitly represent dialect-dependent phonological structure remain limited. This gap motivates our work. We introduce a phonologically grounded, dialect-aware vocabulary representation together with a structured decoding mechanism that jointly models syllable components, enabling Vietnamese ASR systems to better capture systematic pronunciation differences and facilitating computational analysis of dialectal phonetic variation.

\section{Background}
\label{sec:preliminaries}

\begin{figure}[t]
    \centering
    \includegraphics[width=\linewidth]{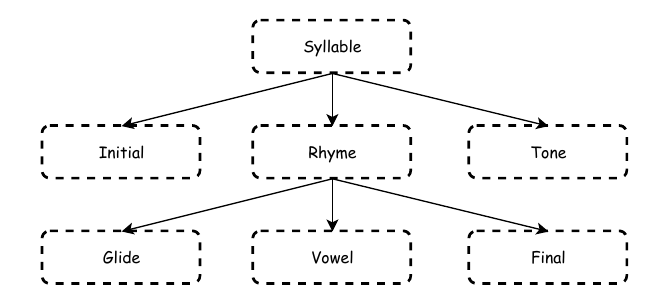}
    \caption{Vietnamese syllable structure.}
    \label{fig:syllabic-structure}
\end{figure}

\subsection{Vietnamese Syllable Structure}
\label{sec:preliminaries-syllable}

Vietnamese is commonly described as a monosyllabic and tonal language in which lexical items correspond to single syllables with a highly regular internal structure~\cite{thuat2016,hao1998,giap2008,giap2011}. Each syllable can be decomposed into three primary phonological components: an \textit{initial} consonant, a \textit{rhyme}, and a \textit{tone}. The rhyme itself may further contain a glide, a vowel nucleus, and an optional final consonant, where the vowel nucleus is obligatory~\cite{thuat2016,hao1998} (Figure \ref{fig:syllabic-structure}). As a result, the rhyme functions as the central carrier of phonological information in Vietnamese syllables.

Vietnamese orthography exhibits a relatively transparent relationship between graphemes and phonemes. The modern writing system, based on the Latin alphabet, was developed to represent Vietnamese speech sounds directly and preserves a strong correspondence between orthographic forms and phonological structure. Consequently, the internal structure of syllables is often reflected explicitly in the written form. 

These linguistic characteristics make Vietnamese particularly suitable for phonologically structured representations. In this work, we exploit this regular syllable structure by decomposing each syllable into its phonological components (initial, rhyme, and tone) and modeling them explicitly within a dialect-aware phonetic vocabulary. Phonemes are denoted using slashes (e.g., /\textipa{i}/), while dialect-specific phones are denoted using square brackets (e.g., [\textipa{1}]). A detailed description of Vietnamese phoneme inventories and orthographic mappings is provided in Appendix~\ref{app:phonetics-orthography}.

\subsection{Phonetic Variation in Vietnamese Multi-Dialect at the Phone Level}
\label{sec:preliminaries-examples}



Vietnamese exhibits substantial dialectal variation across three major dialect regions: Northern, Central, and Southern~\cite{hao1998,chau-dialect}. Although speakers across these regions share the same lexical inventory and orthographic system, many words are realized with systematically different pronunciations. Importantly, most of these differences occur at the phonetic level rather than the lexical level.

A common source of variation arises from interactions between vowels and final consonants within the rhyme. For example, in the Northern dialect, rhymes containing front vowels /\textipa{i, e, E}/ followed by velar consonants /\textipa{k, \ng}/ often undergo articulatory adjustments due to incompatibility between the tongue positions required for these sounds. Front vowels require the tongue to be positioned toward the front of the mouth, whereas velar consonants require contact with the velum. This interaction frequently results in vowel centralization (e.g., /\textipa{i}/ $\rightarrow$ [\textipa{1}]) and consonant palatalization (e.g., /\textipa{k}/ $\rightarrow$ [\textipa{c}]), sometimes accompanied by a transitional glide [\textipa{j}].

Similar phonetic processes occur across other dialect regions, although the exact realizations differ. For instance, Central dialects exhibit multiple province-level variations in how rhymes containing front vowels and velar consonants are realized, while Southern dialects often show diphthong reduction and tone mergers. These systematic differences illustrate that Vietnamese dialect variation largely follows consistent phonetic rules affecting syllable components rather than altering lexical identities.

The structured nature of Vietnamese syllables and the rule-governed character of dialectal variation provide strong motivation for modeling speech at the phonetic component level. A more detailed analysis of phonetic inventories, orthographic correspondences, and dialect-specific pronunciation patterns is provided in Appendix~\ref{app:phonetics-orthography} and Appendix~\ref{app:multi-dialect}.

\section{Methodology}

Our goal is to design a speech recognition framework that explicitly represents Vietnamese phonological structure and dialectal pronunciation patterns. Instead of treating dialect variation purely as acoustic variability, our approach encodes systematic phonetic differences directly in both vocabulary representation and decoding. 

The framework consists of three components: 
(i) a dialect-aware phonetic vocabulary, 
(ii) a structured transcript representation based on syllable components, and 
(iii) a phonetic-structure decoder that jointly predicts these components.

\subsection{Dialect-aware Phonetic Vocabulary}
\label{sec:phonetic-vocab}

Vietnamese syllables follow a regular phonological template consisting of an \emph{initial consonant}, \emph{glide}, \emph{vowel nucleus}, \emph{coda}, and \emph{tone}. Because dialectal variation systematically affects these components, we first apply a deterministic syllable parser that decomposes each orthographic word into its phonological units. The parser enforces Vietnamese phonotactic constraints, including initial–vowel compatibility, coda admissibility, and orthographic exceptions.

Each component is then mapped to its canonical International Phonetic Alphabet (IPA) representation using a \textit{Base Converter} that encodes the standard Vietnamese phonological inventory. These IPA sequences form a dialect-independent reference layer.

\begin{figure}[t]
    \centering
    \includegraphics[width=0.95\linewidth]{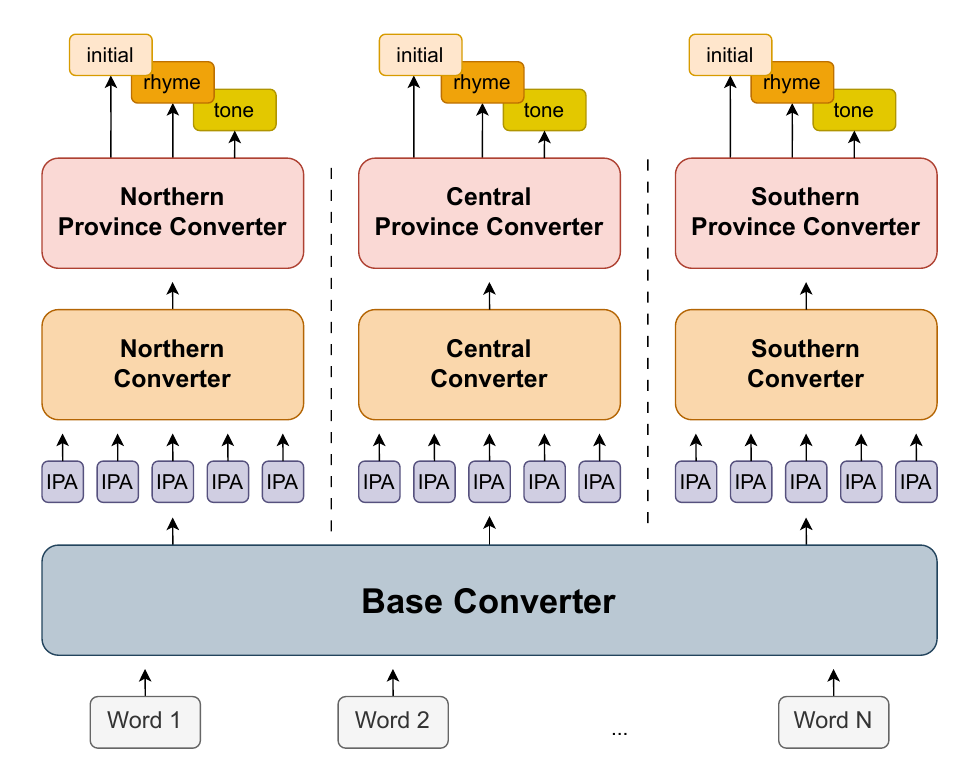}
    \caption{Construction pipeline for the dialect-aware phonetic vocabulary.}
    \label{fig:dialect-converter}
\end{figure}

Dialectal variation is incorporated through a hierarchical conversion pipeline (Figure~\ref{fig:dialect-converter}). A \textit{Dialect Converter} models systematic pronunciation shifts across Northern, Central, and Southern dialects, including consonant mergers, vowel quality changes, diphthong simplification, and coda neutralization. A subsequent \textit{Province Converter} captures selected province-level sociophonetic differences. This modular design separates orthographic representation from phonetic realization while enabling systematic dialect modeling.

\subsection{Multi-dialect Transcript Representation}

Rhymes, comprising the medial, nucleus, and coda, form the phonological core of Vietnamese syllables and encode many dialect-dependent variations~\cite{BenPham_2016}. To balance linguistic expressiveness and decoding efficiency, we adopt a structured representation using three components: \textit{initial}, \textit{rhyme}, and \textit{tone}. 

The vocabulary is therefore partitioned into three disjoint sets: 
$V_{\text{initial}}$ (27 categories), 
$V_{\text{rhyme}}$ (240 categories), 
and $V_{\text{tone}}$ (6 categories). 

During training, each syllable is represented as a triplet $(\text{initial}, \text{rhyme}, \text{tone})$, and utterances are modeled as sequences of these triplets. At inference time, predicted phonetic sequences are deterministically mapped back to orthographic words through a reverse lexicon, leveraging the relatively consistent phonetic–orthographic relationship in Vietnamese.

\subsection{Phonetic-Structure Decoder}

\begin{figure}[t]
    \centering
    \includegraphics[width=0.95\linewidth]{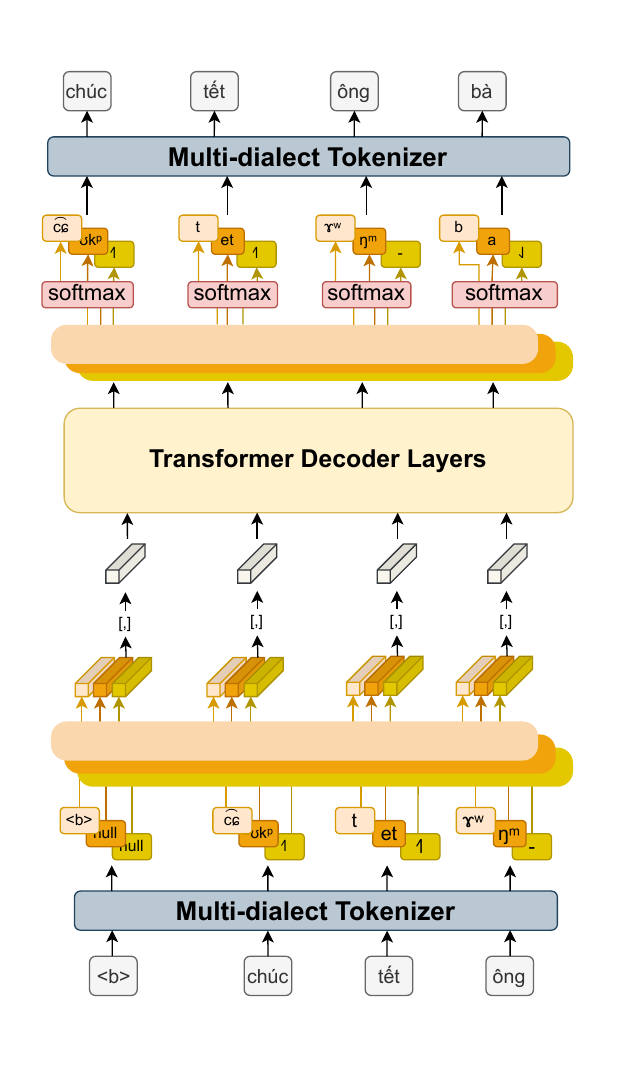}
    \caption{The Phonetic-Structure decoder.}
    \label{fig:syllabic-structure-decoder}
\end{figure}

To exploit syllable structure, we introduce a \emph{multi-token autoregressive decoder} that predicts the initial, rhyme, and tone jointly at each timestep while maintaining autoregressive dependencies across syllables (Figure~\ref{fig:syllabic-structure-decoder}).

\textbf{Component Embeddings.}
At timestep $t$, the decoder receives the previously predicted triplet $(\text{initial}_{t-1}, \text{rhyme}_{t-1}, \text{tone}_{t-1})$. Each component is embedded in a dedicated representation space.

\textbf{Unification Layer.}
The embeddings are combined using a learned unification function that models interactions between syllable components:
\[
h_t = f_{\text{unify}}(e^{\text{initial}}_t,\, e^{\text{rhyme}}_t,\, e^{\text{tone}}_t).
\]

\textbf{Acoustic Attention.}
The unified representation attends to encoder outputs via multi-head attention~\cite{AED_VaswaniSPUJGKP17}, integrating phonetic structure with acoustic context.

\textbf{Parallel Component Prediction.}
A shared projection layer produces a probability vector that is partitioned into three distributions:
\[
p(\text{initial}_t), \quad
p(\text{rhyme}_t), \quad
p(\text{tone}_t).
\]

Joint prediction enforces phonological consistency and reduces invalid syllable combinations.

\subsection{Dialect Analysis Capability}

Because the decoder predicts interpretable phonological components (\textit{initial}, \textit{rhyme}, \textit{tone}), dialect-sensitive pronunciation patterns can be analyzed directly at the sub-syllabic level. This representation enables quantitative comparison of pronunciation across dialect regions and speakers, allowing the model to support both speech recognition and large-scale dialect analysis.

\section{Experiments}
\subsection{Baselines, Datasets, and Metrics}

\paragraph{Baselines.}
We compare the proposed dialect-aware phonetic vocabulary and structured decoding framework with several strong baselines representing common design choices in Vietnamese ASR. 

First, we include \textit{zero-shot} baselines consisting of publicly available large-scale Vietnamese ASR models: PhoWhisper \cite{Pho_LeNN24} and wav2vec~2.0 variants \cite{Wav2vec2_BaevskiZMA20} (wav2vec2-base-vietnamese, wav2vec2-base-vietnamese-160h, and wav2vec2-base-vietnamese-250h). These models serve as off-the-shelf references trained on large external corpora.

For fully supervised comparisons, we implement two widely used encoder--decoder architectures, Speech Transformer and Conformer \cite{sTrans_DongXX18, Conf_GulatiQCPZYHWZW20}, trained with conventional \textit{word-based} tokenization where output units correspond to orthographic Vietnamese words.

To isolate the effect of vocabulary design, we construct \textit{phone-based} variants that replace word tokens with phonetic units (initial, rhyme, tone) derived from our dialect-aware vocabulary, while keeping the architecture, parameter budget, and training settings unchanged.

We further include a multi-task (MT) baseline that jointly predicts dialect labels and transcriptions, representing prior approaches that incorporate dialect information at the acoustic modeling level \cite{Multitask_DanZBWJ22}.

Finally, our full model combines the proposed phonetic vocabulary with the Vietnamese phonetic-structure multi-token decoder, enabling analysis of the contributions from both phonetic representation and structured decoding. This design allows controlled comparisons to evaluate the impact of phonetic vocabulary, structured decoding, and dialect-aware modeling.

\paragraph{Datasets.}
All experiments are conducted on the publicly available Multi-Dialect Vietnamese (UIT-ViMD) corpus \cite{ViMD_DinhDNN24}, which contains speech collected from speakers across all Vietnamese provinces. We follow the official train--validation--test splits, ensuring speaker disjointness while preserving the original dialectal distribution. Only minimal preprocessing is applied: non-Vietnamese tokens are treated as noise and removed from the transcript, and audio is kept in its original mono 16\,kHz format to preserve natural pronunciation and recording conditions. UIT-ViMD therefore provides a challenging benchmark for evaluating ASR robustness across Vietnamese dialects. While the UIT-ViMD corpus is moderate in size, it provides balanced dialectal coverage across Vietnamese regions, making it suitable for evaluating dialect-aware modeling approaches.

\paragraph{Metrics.}
We evaluate models from both lexical and phonetic perspectives. Word Error Rate (WER) is reported as the primary metric, together with dialect-wise results. Provinces in Northern Vietnam are grouped as Northern dialect, provinces from Nghệ An–Hà Tĩnh to Thừa Thiên Huế as Central dialect, and the remaining provinces as Southern dialect.

For phone-based systems, phonetic accuracy is measured using Phone Error Rate (PER) on predicted phonetic sequences. We additionally report component-wise error rates for initials, rhymes, and tones to analyze phonological modeling behavior. These metrics jointly capture transcription accuracy, phonetic fidelity, and dialect robustness.

\subsection{Configuration}

\begin{table*}[ht]
\centering
\small
\resizebox{\textwidth}{!}{
\setlength{\tabcolsep}{5pt}
\begin{tabular}{l c r cc ccc c}
\toprule
\textbf{Model} &
\textbf{Setup} &
\textbf{Params} &
\textbf{WER}$\downarrow$ &
\textbf{PER}$\downarrow$ &
\textbf{WER$_\text{N}$} &
\textbf{WER$_\text{C}$} &
\textbf{WER$_\text{S}$} &
\textbf{Time(s)}$\downarrow$ \\
\midrule

\multicolumn{9}{l}{\textit{Published Zero-shot Baselines$^\dagger$}} \\
\midrule
PhoWhisper-base       & -- & 74M  & 18.61 & -- & 14.69 & 24.15 & 17.87 & 0.73 \\
Whisper-base          & -- & 244M & 31.38 & -- & 26.37 & 39.91 & 29.46 & 1.21 \\
wav2vec2-base-vi      & -- & 95M  & 23.07 & -- & 20.32 & 27.28 & 22.48 & 0.03 \\
wav2vec2-base-vi-160h & -- & 95M  & 31.74 & -- & 27.50 & 38.12 & 30.93 & 0.03 \\
wav2vec2-base-vi-250h & -- & 95M  & 17.47 & -- & 14.98 & 20.97 & 17.24 & 0.07 \\

\midrule
\multicolumn{9}{l}{\textit{Published Fine-tuned Baselines$^{\dagger}$}} \\
\midrule
PhoWhisper-base        & FT & 74M & 16.30 & -- & 13.20 & 18.26 & \textbf{13.54} & -- \\
Whisper-base           & FT & 74M & 19.93 & -- & 20.05 & 27.89 & 20.89 & -- \\
wav2vec2-base-vi       & FT & 95M & 15.80 & -- & 14.64 & 20.27 & 17.72 & -- \\
wav2vec2-base-vi-160h  & FT & 95M & 17.49 & -- & 16.70 & 24.56 & 20.51 & -- \\
wav2vec2-base-vi-250h  & FT & 95M & 13.56 & -- & 12.29 & 17.15 & 15.26 & -- \\

\midrule
\multicolumn{9}{l}{\textit{Our Implementations}} \\
\midrule
Transformer            & --   & 29M & 16.71 $\pm$ 0.13 & --   & 14.32 $\pm$ 0.07 & 18.24 $\pm$ 0.14 & 17.49 $\pm$ 0.06 & 0.37 \\
Conformer              & --   & 31M & 17.49 $\pm$ 0.08 & --   & 16.71 $\pm$ 0.06 & 19.68 $\pm$ 0.11 & 18.52 $\pm$ 0.03 & 0.29 \\
Transformer            & P    & 26M & 15.53 $\pm$ 0.08 & 12.47 $\pm$ 0.16 & 13.31 $\pm$ 0.13 & 15.77 $\pm$ 0.08 & 16.46 $\pm$ 0.01 & 1.25 \\
Conformer              & P    & 28M & 13.89 $\pm$ 0.11 & 11.22 $\pm$ 0.09 & 12.51 $\pm$ 0.13 & 17.52 $\pm$ 0.10 & 15.22 $\pm$ 0.07 & 0.93 \\
MT-Transformer         & P+M  & 26M & 15.90 $\pm$ 0.10 & 12.95 $\pm$ 0.07 & 14.27 $\pm$ 0.14 & 17.56 $\pm$ 0.07 & 17.15 $\pm$ 0.09 & 1.58 \\
MT-Conformer           & P+M  & 28M & 16.77 $\pm$ 0.19 & 13.50 $\pm$ 0.13 & 14.63 $\pm$ 0.17 & 19.53 $\pm$ 0.04 & 17.71 $\pm$ 0.13 & 1.21 \\
\midrule
\textbf{Our-Transformer} & \textbf{P+V} & 26M & \textbf{13.05 $\pm$ 0.09} & \textbf{8.36 $\pm$ 0.12} & \textbf{11.35 $\pm$ 0.10} & \textbf{15.37 $\pm$ 0.15} & 14.05 $\pm$ 0.06 & 0.33 \\
Our-Conformer          & P+V  & 28M & 13.27 $\pm$ 0.16 & 8.47 $\pm$ 0.14 & 11.29 $\pm$ 0.21 & 16.21 $\pm$ 0.18 & 14.08 $\pm$ 0.03 & 0.28 \\

\bottomrule
\end{tabular}}
\caption{Performance comparison on Vietnamese multi-dialect ASR.
Metrics include word error rate (WER, \%) and phone error rate (PER, \%).
WER$_\text{N}$, WER$_\text{C}$, and WER$_\text{S}$ denote Northern, Central, and Southern dialects.
Setup: phonetic supervision (P), multi-task learning (M), Vietnamese phonetic-structure decoder (V).
$^\dagger$ Results reported from \cite{ViMD_DinhDNN24}. Note that our implementations are four-time run while published results are single run.}
\label{tab:main_eval_expanded}
\end{table*}

All supervised models are trained from scratch on the UIT-ViMD corpus without external language models using a single NVIDIA H100 GPU. Experiments are implemented with the SpeechBrain end-to-end speech processing toolkit~\cite{SpeechBrain}. The reported results are the average values of four-time training.

Speech signals are represented using 80-dimensional log-Mel filterbank (Fbank) features with a 25\,ms window and a 10\,ms frame shift. SpecAugment~\cite{SpecAug_ParkCZCZCL19} is applied during training using both time and frequency masking.

For encoder--decoder architectures, the Transformer uses 12 encoder layers and 6 decoder layers. The Conformer model consists of 8 encoder layers and 4 decoder layers with an attention dimension of 256, 4 attention heads, and a feed-forward dimension of 2048.

Models are optimized using Adam~\cite{Adam_KingmaB14} with the Noam learning-rate schedule. Warm-up steps~\cite{Warmup_GotmareKXS19} are set to 40k for Transformer and 20k for Conformer, with initial learning rates of $10^{-3}$ and $4\times10^{-4}$, respectively. Label smoothing~\cite{LabelSmooth_SzegedyVISW16} and dropout are both set to 0.1.

Training employs a joint CTC--Attention objective~\cite{JointCTCAED_KimHW17} with CTC weights of 0.30 for Transformer and 0.15 for Conformer, which encourages monotonic alignment while preserving the flexibility of attention-based decoding.

\subsection{Main Results}

\begin{figure*}[t]
    \centering
    \includegraphics[width=0.95\textwidth]{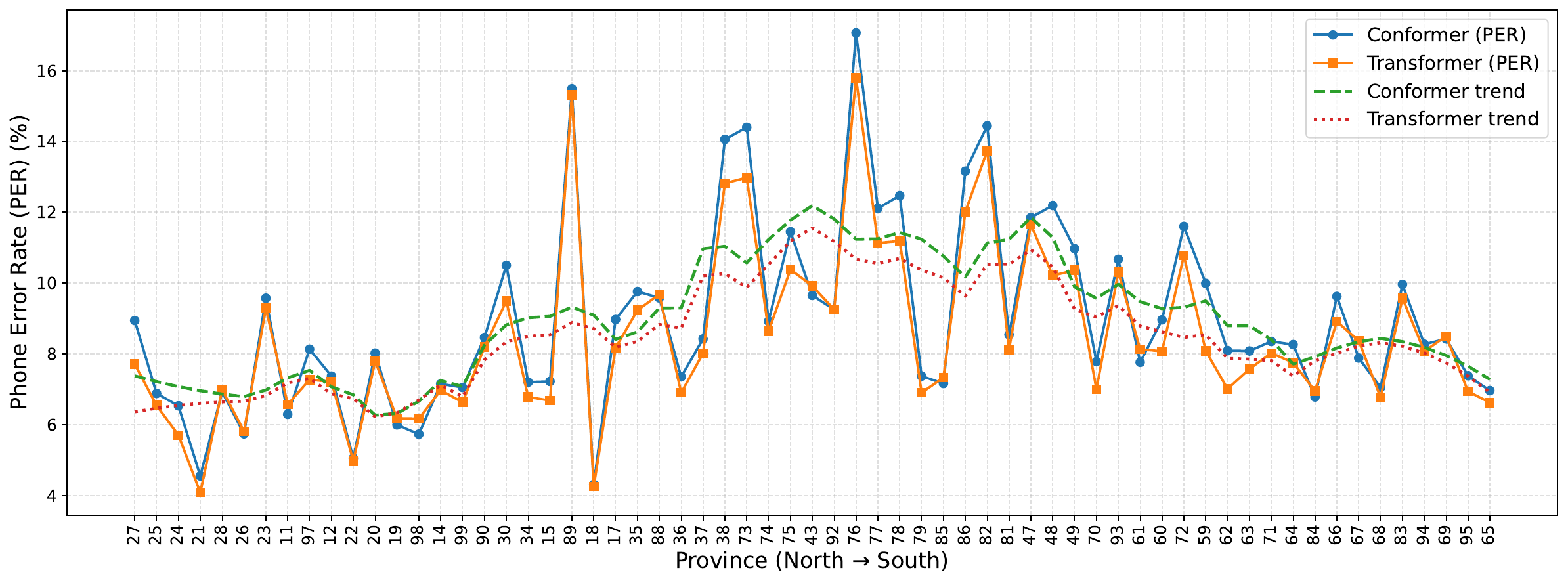}
    \caption{Phone error rate (PER, \%) by province for Transformer and Conformer models using the proposed Vietnamese phonetics method. The reported PER is the best one among four-time training.}
    \label{fig:per_by_prov}
\end{figure*}

Table~\ref{tab:main_eval_expanded} reports the main ASR results on the UIT-ViMD benchmark, comparing the proposed dialect-aware phonetic framework with strong zero-shot, fine-tuned, and supervised baselines. Overall, the results show that explicitly modeling Vietnamese phonological structure and dialectal variation improves recognition accuracy across architectures and dialect regions.

Large pretrained models such as PhoWhisper and wav2vec~2.0 achieve strong performance by leveraging large-scale pretraining. However, our models trained from scratch on UIT-ViMD achieve better WERs with substantially fewer parameters. This result suggests that incorporating linguistic structure---specifically phonetic and dialect-aware representations---can partially compensate for the absence of large-scale pretraining in dialect-rich settings.

\textbf{Effect of phonetic vocabulary design.}
Replacing word-level outputs with the proposed phonetic units consistently improves recognition performance. For instance, Transformer-P reduces WER from 16.71\% to 15.53\%, while Conformer-P achieves a larger reduction from 17.49\% to 13.89\%. These improvements indicate that phonetic representations better align acoustic signals with linguistic units, reducing acoustic–lexical mismatch across dialects.

\textbf{Impact of the phonetic-structure decoder.}
The full phonetic framework further improves performance when combined with the Vietnamese phonetic-structure decoder. Our-Transformer achieves the best overall result with a WER of 13.05\% and a PER of 8.36\%, while maintaining a compact model size. This result improves upon the best fine-tuned wav2vec2 baseline (13.56\%) while using a significantly smaller model and no external pretraining. By jointly predicting initials, rhymes, and tones, the decoder enforces phonological consistency and reduces invalid syllable hypotheses, resulting in more stable decoding. These results suggest that phonological structure provides a robust inductive bias for modeling systematic dialect variation.

Dialect-wise evaluation shows consistent gains across Northern, Central, and Southern dialect groups. Improvements are particularly noticeable in Central dialect provinces, which exhibit greater pronunciation variability. Figure~\ref{fig:per_by_prov} further illustrates this trend, where phone error rates remain lower and more stable across provinces when using the proposed phonetic framework.

\subsection{Analysis of Results}

\begin{table*}[t] 
    \centering 
    \small 
    \setlength{\tabcolsep}{4pt} 
    \begin{tabular}{l cccc cccc} \toprule 
    \multirow{2}{*}{\textbf{Design}} & \multicolumn{4}{c}{\textbf{Transformer}} & \multicolumn{4}{c}{\textbf{Conformer}} \\ 
    \cmidrule(lr){2-5} \cmidrule(lr){6-9} & \textbf{PER}$\downarrow$ & Initial & Rhyme & Tone & \textbf{PER}$\downarrow$ & Initial & Rhyme & Tone \\ 
    \midrule Phonetic baseline & 12.52 & 12.29 & 14.53 & 10.52 & 11.29 & 11.07 & 13.33 & 9.26 \\ 
    + Multitask & 12.99 & 12.66 & 15.15 & 10.90 & 13.58 & 13.01 & 15.94 & 11.47 \\ 
    + \textbf{Phonetic-structure decoder (ours)} & \textbf{8.42} & \textbf{8.36} & \textbf{10.07} & \textbf{6.88} & 8.54 & 8.82 & 10.63 & 7.24 \\ 
    \bottomrule 
    \end{tabular} 
    \caption{Phonetic error analysis of phone-based ASR models. Metrics include overall phone error rate (PER, \%) and component-wise PERs for initial, rhyme, and tone units. The reported results are the best one among four-time training.}
    \label{tab:component_eval} 
\end{table*}

\begin{table*}[t]
    \centering
    \small
    \setlength{\tabcolsep}{5pt}
    \begin{tabular}{l l c c c c}
    \toprule
    \textbf{Modeling Unit}
    & \textbf{Architecture}
    & \textbf{Unique Correct Words}$\uparrow$
    & \textbf{Pearson $r$}
    & \textbf{Spearman $\rho$} \\
    \midrule
    \multirow{2}{*}{Word-level}
    & Transformer & 1{,}696 & 0.79 & 0.76 \\
    & Conformer   & 1{,}718 & 0.77 & 0.74 \\
    \midrule
    \multirow{2}{*}{\textbf{Phonetic-level}}
    & Transformer & \textbf{1{,}950} & 0.57 & 0.45 \\
    & Conformer   & \textbf{1{,}943} & 0.58 & 0.47 \\
    \bottomrule
    \end{tabular}
    \caption{
    Lexical diversity and frequency bias analysis for word-level and phonetic-level ASR models. Metrics include the number of unique word types correctly recognized at least once in the test set, as well as Pearson/Spearman correlations between word frequency in the training data and per-word recall.
}
\label{tab:lexical_diversity}
\end{table*}

\paragraph{Phonetic Structure and Dialect Modeling}

Table~\ref{tab:component_eval} provides a component-wise analysis of phonetic prediction errors. Although phonetic baselines already outperform word-based systems, they still exhibit relatively high error rates for initials, rhymes, and tones. This reflects the ambiguity that arises when these components are predicted independently without structural constraints.

Introducing the phonetic-structure decoder substantially reduces error rates across all components. For Transformer, overall PER decreases from 12.52\% to 8.42\%, while Conformer improves from 11.29\% to 8.54\%. Improvements are consistent for initials, rhymes, and tones, indicating that jointly modeling syllable components helps enforce phonological consistency during decoding.

By comparison, multi-task learning with dialect classification improves dialect identification accuracy (Table~\ref{tab:detect_eval}) but does not consistently improve ASR performance. In some cases, PER slightly degrades relative to phonetic-only baselines. This suggests that incorporating dialect information solely through acoustic supervision may introduce competing optimization objectives, particularly when training data is limited. In contrast, our approach integrates dialectal knowledge directly into the phonetic vocabulary and decoding process, leading to more stable performance gains.

\paragraph{Word-level vs. Phonetic-level Modeling}

Table~\ref{tab:lexical_diversity} compares lexical diversity and frequency bias between word-level and phonetic-level ASR systems. Phonetic-level models correctly recognize a substantially larger number of unique word types in the test set (1,950 vs.\ 1,696 for Transformer and 1,943 vs.\ 1,718 for Conformer), indicating improved generalization to infrequent lexical items.

Furthermore, correlations between training-set word frequency and test-time recall are significantly lower for phonetic-level models. Both Pearson and Spearman coefficients decrease notably, suggesting that phonetic representations reduce reliance on frequency-driven memorization. Instead, the model generalizes through shared phonological structure across syllables. This behavior is particularly beneficial for Vietnamese, where systematic pronunciation variation across dialects often affects phonological components rather than entire lexical items.

\paragraph{Dialect Classification Performance}

\begin{table}[ht] 
    \centering 
    \resizebox{0.45\textwidth}{!}{ 
    \begin{tabular}{lcc} \hline 
    \textbf{Model} & \textbf{Accuracy} & \textbf{Macro-F1} \\ \hline 
    \multicolumn{3}{l}{\textit{Published baselines}$^\dagger$} \\ \hline 
    PhoWhisper-base & - & 86.97 \\ 
    Whisper-base & - & 85.59 \\ 
    wav2vec2-base-vi & - & 91.02 \\ 
    wav2vec2-large-vi & - & 91.47 \\ 
    wav2vec2-xlsr-300m & - & 89.01 \\
    wav2vec2-large-xlsr-35 & - & 87.36 \\ \hline 
    \multicolumn{3}{l}{\textit{Our implementations}} \\ \hline 
    \textbf{Our-Transformer} & \textbf{95.75 $\pm$ 0.11} & \textbf{96.11 $\pm$ 0.09} \\ 
    Our-Conformer & 95.47 $\pm$ 0.11 & 95.80 $\pm$ 0.13 \\ \hline 
    \end{tabular} } 
    \caption{Classification performance (\%) of multi-task phonetic models. Metrics include macro F1 and accuracy. $^{\dagger}$Results reported from \cite{ViMD_DinhDNN24}.} 
    \label{tab:detect_eval} 
\end{table}

Table~\ref{tab:detect_eval} reports dialect classification performance for the proposed multi-task phonetic models. Both Transformer and Conformer achieve high accuracy (95.75\% and 95.47\%) and macro-F1 scores (96.11\% and 95.80\%), matching or exceeding previously reported baselines~\cite{ViMD_DinhDNN24}.

These results indicate that the learned phonetic representations capture strong dialect-discriminative signals. Importantly, this information emerges naturally from phonological modeling rather than being explicitly engineered for dialect classification. The findings therefore suggest that phonologically structured ASR models can simultaneously support speech recognition and large-scale analysis of dialect variation, providing a useful computational framework for studying Vietnamese sociophonetic diversity. This capability highlights the potential of phonologically informed ASR systems as tools for computational studies of dialect variation and regional language change.

\section{Conclusion}

This paper addresses a key limitation in Vietnamese multi-dialect ASR: the common assumption that orthography corresponds uniformly to pronunciation across dialects. Motivated by linguistic evidence that Vietnamese dialectal variation is systematic and largely realized at the phonetic level, we propose a dialect-aware phonetic framework that integrates phonological structure into both vocabulary design and decoding. Experiments on the UIT-ViMD benchmark show that incorporating phonological structure improves recognition accuracy and reduces frequency-driven lexical bias. Importantly, the proposed representation provides interpretable phonological signals that capture systematic pronunciation differences across dialect regions. These findings highlight the potential of linguistically informed ASR systems not only for robust speech recognition but also for computational analysis of dialect variation and sociophonetic patterns. 



\section*{Limitations}

While the proposed framework demonstrates significant results for Vietnamese multi-dialect ASR, several limitations remain. The phonetic-structure decoder models syllable-level components (initial, rhyme, and tone), and implicitly captures the cross-syllabic or cross-word phonological and prosodic phenomena, which may also influence dialectal pronunciation patterns, by attention mechanism. Moreover, as discussed in Appendix \ref{app:multi-dialect}, converting phonetic forms back into orthographic forms is ambiguous for some vowels and initials among dialects. Incorporating an additional language model for this conversion could yield promising improvements. However, in this study, we intentionally keep the Phonetic-Structure Decoder standalone in order to evaluate the intrinsic capability of the proposed method. Realistic implementation can consider having additional modules as a language model for yielding better scores.

Our future works will consider the explicit modeling of these factors for better multi-dialect speech description. In addition, our experiments focus on models trained primarily on the UIT-ViMD corpus. Future work could explore integrating the proposed phonetic representation with larger pretrained acoustic models and evaluating the framework on broader multilingual and multi-dialect speech datasets.

\bibliography{acl_latex}

@inproceedings{ViMD_DinhDNN24,
    author       = {Nguyen Dinh and
                  Thanh Dang and
                  Luan Thanh Nguyen and
                  Kiet Van Nguyen},
    editor       = {Yaser Al{-}Onaizan and
                  Mohit Bansal and
                  Yun{-}Nung Chen},
    title        = {Multi-Dialect Vietnamese: Task, Dataset, Baseline Models and Challenges},
    booktitle    = {Proceedings of the 2024 Conference on Empirical Methods in Natural
                  Language Processing, {EMNLP} 2024, Miami, FL, USA, November 12-16,
                  2024},
    pages        = {7476--7498},
    publisher    = {Association for Computational Linguistics},
    year         = {2024},
    url          = {https://doi.org/10.18653/v1/2024.emnlp-main.426},
    doi          = {10.18653/V1/2024.EMNLP-MAIN.426},
    bibsource    = {dblp computer science bibliography, https://dblp.org}
}

@article{BenPham_2016,
    author  = {Ben Phạm and Sharynne McLeod},
    title   = {Consonants, vowels and tones across Vietnamese dialects},
    journal = {International Journal of Speech-Language Pathology},
    year    = {2016},
    volume  = {18},
    number  = {2},
    pages   = {122--134},
}

@article{Multitask_DanZBWJ22,
    author       = {Zhengjia Dan and
                  Yue Zhao and
                  Xiaojun Bi and
                  Licheng Wu and
                  Qiang Ji},
    title        = {Multi-Task Transformer with Adaptive Cross-Entropy Loss for Multi-Dialect
                  Speech Recognition},
    journal      = {Entropy},
    volume       = {24},
    number       = {10},
    pages        = {1429},
    year         = {2022},
    url          = {https://doi.org/10.3390/e24101429},
    doi          = {10.3390/E24101429},
    bibsource    = {dblp computer science bibliography, https://dblp.org}
}

@inproceedings{Multitask_ElfekyBVMW16,
    author       = {Mohamed Elfeky and
                  Meysam Bastani and
                  Xavier Velez and
                  Pedro J. Moreno and
                  Austin Waters},
    title        = {Towards acoustic model unification across dialects},
    booktitle    = {2016 {IEEE} Spoken Language Technology Workshop, {SLT} 2016, San Diego,
                  CA, USA, December 13-16, 2016},
    pages        = {624--628},
    publisher    = {{IEEE}},
    year         = {2016},
    url          = {https://doi.org/10.1109/SLT.2016.7846328},
    doi          = {10.1109/SLT.2016.7846328},
    bibsource    = {dblp computer science bibliography, https://dblp.org}
}

@article{Transfer_TaLD24,
    author       = {Bao Thang Ta and
                  Nhat Minh Le and
                  Van Hai Do},
    title        = {Transfer learning methods for low-resource speech accent recognition:
                  {A} case study on Vietnamese language},
    journal      = {Eng. Appl. Artif. Intell.},
    volume       = {132},
    pages        = {107895},
    year         = {2024},
    url          = {https://doi.org/10.1016/j.engappai.2024.107895},
    doi          = {10.1016/J.ENGAPPAI.2024.107895},
    bibsource    = {dblp computer science bibliography, https://dblp.org}
}

@inproceedings{Transfer_SuwanbanditNSC23,
    author       = {Artit Suwanbandit and
                  Burin Naowarat and
                  Orathai Sangpetch and
                  Ekapol Chuangsuwanich},
    editor       = {Naomi Harte and
                  Julie Carson{-}Berndsen and
                  Gareth Jones},
    title        = {Thai Dialect Corpus and Transfer-based Curriculum Learning Investigation
                  for Dialect Automatic Speech Recognition},
    booktitle    = {24th Annual Conference of the International Speech Communication Association,
                  Interspeech 2023, Dublin, Ireland, August 20-24, 2023},
    pages        = {4069--4073},
    publisher    = {{ISCA}},
    year         = {2023},
    url          = {https://doi.org/10.21437/Interspeech.2023-1828},
    doi          = {10.21437/INTERSPEECH.2023-1828},
    bibsource    = {dblp computer science bibliography, https://dblp.org}
}

@inproceedings{SpecAug_ParkCZCZCL19,
    author       = {Daniel S. Park and
                  William Chan and
                  Yu Zhang and
                  Chung{-}Cheng Chiu and
                  Barret Zoph and
                  Ekin D. Cubuk and
                  Quoc V. Le},
    editor       = {Gernot Kubin and
                  Zdravko Kacic},
    title        = {SpecAugment: {A} Simple Data Augmentation Method for Automatic Speech
                  Recognition},
    booktitle    = {20th Annual Conference of the International Speech Communication Association,
                  Interspeech 2019, Graz, Austria, September 15-19, 2019},
    pages        = {2613--2617},
    publisher    = {{ISCA}},
    year         = {2019},
    url          = {https://doi.org/10.21437/Interspeech.2019-2680},
    doi          = {10.21437/INTERSPEECH.2019-2680},
    bibsource    = {dblp computer science bibliography, https://dblp.org}
}

@inproceedings{Aformer_WangLLW23,
  author       = {Xuefei Wang and
                  Yanhua Long and
                  Yijie Li and
                  Haoran Wei},
  editor       = {Naomi Harte and
                  Julie Carson{-}Berndsen and
                  Gareth Jones},
  title        = {Multi-pass Training and Cross-information Fusion for Low-resource
                  End-to-end Accented Speech Recognition},
  booktitle    = {24th Annual Conference of the International Speech Communication Association,
                  Interspeech 2023, Dublin, Ireland, August 20-24, 2023},
  pages        = {2923--2927},
  publisher    = {{ISCA}},
  year         = {2023},
  url          = {https://doi.org/10.21437/Interspeech.2023-142},
  doi          = {10.21437/INTERSPEECH.2023-142},
  bibsource    = {dblp computer science bibliography, https://dblp.org}
}

@inproceedings{SurveyVNASR_Nga2021,
    author={Nga, Cao Hong and Li, Chung-Ting and Li, Yung-Hui and Wang, Jia-Ching},
    booktitle={2021 9th International Conference on Orange Technology (ICOT)}, 
    title={A Survey of Vietnamese Automatic Speech Recognition}, 
    year={2021},
    volume={},
    number={},
    pages={1-4},
}

@inproceedings{MoE_ZhouGYDW24,
    author       = {Jie Zhou and
                  Shengxiang Gao and
                  Zhengtao Yu and
                  Ling Dong and
                  Wenjun Wang},
    editor       = {Maosong Sun and
                  Jiye Liang and
                  Xianpei Han and
                  Zhiyuan Liu and
                  Yulan He and
                  Gaoqi Rao and
                  Yubo Chen and
                  Zhiliang Tian},
    title        = {DialectMoE: An End-to-End Multi-dialect Speech Recognition Model with
                  Mixture-of-Experts},
    booktitle    = {Chinese Computational Linguistics - 23rd China National Conference,
                  {CCL} 2024, Taiyuan, China, July 25-28, 2024, Proceedings},
    series       = {Lecture Notes in Computer Science},
    volume       = {14761},
    pages        = {243--258},
    publisher    = {Springer},
    year         = {2024},
    url          = {https://doi.org/10.1007/978-981-97-8367-0\_15},
    doi          = {10.1007/978-981-97-8367-0\_15},
    bibsource    = {dblp computer science bibliography, https://dblp.org}
}

@article{VietASR_Tran2024,
AUTHOR = {Tran, Linh Thi Thuc and Kim, Han-Gyu and La, Hoang Minh and Van Pham, Su},
TITLE = {Automatic Speech Recognition of Vietnamese for a New Large-Scale Corpus},
JOURNAL = {Electronics},
VOLUME = {13},
YEAR = {2024},
NUMBER = {5},
ARTICLE-NUMBER = {977},
ISSN = {2079-9292},
}

@inproceedings{AED_VaswaniSPUJGKP17,
    author       = {Ashish Vaswani and
                  Noam Shazeer and
                  Niki Parmar and
                  Jakob Uszkoreit and
                  Llion Jones and
                  Aidan N. Gomez and
                  Lukasz Kaiser and
                  Illia Polosukhin},
    editor       = {Isabelle Guyon and
                  Ulrike von Luxburg and
                  Samy Bengio and
                  Hanna M. Wallach and
                  Rob Fergus and
                  S. V. N. Vishwanathan and
                  Roman Garnett},
    title        = {Attention is All you Need},
    booktitle    = {Advances in Neural Information Processing Systems 30: Annual Conference
                  on Neural Information Processing Systems 2017, December 4-9, 2017,
                  Long Beach, CA, {USA}},
    pages        = {5998--6008},
    year         = {2017},
    url          = {https://proceedings.neurips.cc/paper/2017/hash/3f5ee243547dee91fbd053c1c4a845aa-Abstract.html},
    bibsource    = {dblp computer science bibliography, https://dblp.org}
}

@inproceedings{Pho_LeNN24,
    author       = {Thanh{-}Thien Le and
                  Linh The Nguyen and
                  Dat Quoc Nguyen},
    title        = {PhoWhisper: Automatic Speech Recognition for Vietnamese},
    booktitle    = {The Second Tiny Papers Track at {ICLR} 2024, Tiny Papers @ {ICLR}
                  2024, Vienna, Austria, May 11, 2024},
    publisher    = {OpenReview.net},
    year         = {2024},
    url          = {https://openreview.net/forum?id=x3c3MkJfpG},
    bibsource    = {dblp computer science bibliography, https://dblp.org}
}

@inproceedings{Wav2vec2_BaevskiZMA20,
    author       = {Alexei Baevski and
              Yuhao Zhou and
              Abdelrahman Mohamed and
              Michael Auli},
    editor       = {Hugo Larochelle and
              Marc'Aurelio Ranzato and
              Raia Hadsell and
              Maria{-}Florina Balcan and
              Hsuan{-}Tien Lin},
    title        = {wav2vec 2.0: {A} Framework for Self-Supervised Learning of Speech
              Representations},
    booktitle    = {Advances in Neural Information Processing Systems 33: Annual Conference
              on Neural Information Processing Systems 2020, NeurIPS 2020, December
              6-12, 2020, virtual},
    year         = {2020},
    url          = {https://proceedings.neurips.cc/paper/2020/hash/92d1e1eb1cd6f9fba3227870bb6d7f07-Abstract.html},
    bibsource    = {dblp computer science bibliography, https://dblp.org}
}

@inproceedings{sTrans_DongXX18,
    author       = {Linhao Dong and
                  Shuang Xu and
                  Bo Xu},
    title        = {Speech-Transformer: {A} No-Recurrence Sequence-to-Sequence Model for
                  Speech Recognition},
    booktitle    = {2018 {IEEE} International Conference on Acoustics, Speech and Signal
                  Processing, {ICASSP} 2018, Calgary, AB, Canada, April 15-20, 2018},
    pages        = {5884--5888},
    publisher    = {{IEEE}},
    year         = {2018},
    url          = {https://doi.org/10.1109/ICASSP.2018.8462506},
    doi          = {10.1109/ICASSP.2018.8462506},
    bibsource    = {dblp computer science bibliography, https://dblp.org}
}

@inproceedings{Conf_GulatiQCPZYHWZW20,
    author       = {Anmol Gulati and
                  James Qin and
                  Chung{-}Cheng Chiu and
                  Niki Parmar and
                  Yu Zhang and
                  Jiahui Yu and
                  Wei Han and
                  Shibo Wang and
                  Zhengdong Zhang and
                  Yonghui Wu and
                  Ruoming Pang},
    editor       = {Helen Meng and
                  Bo Xu and
                  Thomas Fang Zheng},
    title        = {Conformer: Convolution-augmented Transformer for Speech Recognition},
    booktitle    = {21st Annual Conference of the International Speech Communication Association,
                  Interspeech 2020, Virtual Event, Shanghai, China, October 25-29, 2020},
    pages        = {5036--5040},
    publisher    = {{ISCA}},
    year         = {2020},
    url          = {https://doi.org/10.21437/Interspeech.2020-3015},
    doi          = {10.21437/INTERSPEECH.2020-3015},
    bibsource    = {dblp computer science bibliography, https://dblp.org}
}

@article{SpeechBrain,
author       = {Mirco Ravanelli and
              Titouan Parcollet and
              Peter Plantinga and
              Aku Rouhe and
              Samuele Cornell and
              Loren Lugosch and
              Cem Subakan and
              Nauman Dawalatabad and
              Abdelwahab Heba and
              Jianyuan Zhong and
              Ju{-}Chieh Chou and
              Sung{-}Lin Yeh and
              Szu{-}Wei Fu and
              Chien{-}Feng Liao and
              Elena Rastorgueva and
              Fran{\c{c}}ois Grondin and
              William Aris and
              Hwidong Na and
              Yan Gao and
              Renato De Mori and
              Yoshua Bengio},
title        = {SpeechBrain: {A} General-Purpose Speech Toolkit},
volume       = {abs/2106.04624},
year         = {2021},
eprinttype    = {arXiv},
}

@inproceedings{Adam_KingmaB14,
    author       = {Diederik P. Kingma and
                  Jimmy Ba},
    editor       = {Yoshua Bengio and
                  Yann LeCun},
    title        = {Adam: {A} Method for Stochastic Optimization},
    booktitle    = {3rd International Conference on Learning Representations, {ICLR} 2015,
                  San Diego, CA, USA, May 7-9, 2015, Conference Track Proceedings},
    year         = {2015},
    url          = {http://arxiv.org/abs/1412.6980},
    bibsource    = {dblp computer science bibliography, https://dblp.org}
}

@inproceedings{Warmup_GotmareKXS19,
    author       = {Akhilesh Gotmare and
                  Nitish Shirish Keskar and
                  Caiming Xiong and
                  Richard Socher},
    title        = {A Closer Look at Deep Learning Heuristics: Learning rate restarts,
                  Warmup and Distillation},
    booktitle    = {7th International Conference on Learning Representations, {ICLR} 2019,
                  New Orleans, LA, USA, May 6-9, 2019},
    publisher    = {OpenReview.net},
    year         = {2019},
    url          = {https://openreview.net/forum?id=r14EOsCqKX},
    bibsource    = {dblp computer science bibliography, https://dblp.org}
}

@inproceedings{LabelSmooth_SzegedyVISW16,
    author       = {Christian Szegedy and
                  Vincent Vanhoucke and
                  Sergey Ioffe and
                  Jonathon Shlens and
                  Zbigniew Wojna},
    title        = {Rethinking the Inception Architecture for Computer Vision},
    booktitle    = {2016 {IEEE} Conference on Computer Vision and Pattern Recognition,
                  {CVPR} 2016, Las Vegas, NV, USA, June 27-30, 2016},
    pages        = {2818--2826},
    publisher    = {{IEEE} Computer Society},
    year         = {2016},
    url          = {https://doi.org/10.1109/CVPR.2016.308},
    doi          = {10.1109/CVPR.2016.308},
    bibsource    = {dblp computer science bibliography, https://dblp.org}
}

@inproceedings{JointCTCAED_KimHW17,
    author       = {Suyoun Kim and
                  Takaaki Hori and
                  Shinji Watanabe},
    title        = {Joint CTC-attention based end-to-end speech recognition using multi-task
                  learning},
    booktitle    = {2017 {IEEE} International Conference on Acoustics, Speech and Signal
                  Processing, {ICASSP} 2017, New Orleans, LA, USA, March 5-9, 2017},
    pages        = {4835--4839},
    publisher    = {{IEEE}},
    year         = {2017},
    url          = {https://doi.org/10.1109/ICASSP.2017.7953075},
    doi          = {10.1109/ICASSP.2017.7953075},
    bibsource    = {dblp computer science bibliography, https://dblp.org}
}

@book{hao1998,
  title={Tiếng Việt mấy vấn đề ngữ âm - ngữ pháp - ngữ nghĩa},
  author={Cao Xuân Hạo},
  year={1998},
  publisher={Nhà xuất bản Giáo dục Việt Nam}
}

@book{giap2011,
    author = {Nguyễn Thiện Giáp},
    title = {Vấn đề "từ" trong tiếng Việt},
    publisher = {Nhà xuất bản Giáo dục Việt Nam},
    year = {2011}
}

@book{giap2008,
    author = {Nguyễn Thiện Giáp},
    title = {Từ vựng học tiếng Việt},
    publisher = {Nhà xuất bản Giáo dục Việt Nam},
    year = {2008}
}

@book{thuat2016,
    author = {Đoàn Thiện Thuật},
    title = {Ngữ âm tiếng Việt},
    publisher = {Nhà xuất bản Đại học Quốc gia Hà Nội},
    year = {2016}
}

@book{chau-dialect,
    author = {Hoàng Thị Châu},
    title = {Phương ngữ học tiếng Việt},
    publisher = {Nhà xuất bản Đại học Quốc gia Hà Nội},
    year = {2002}
}

@book{phe-dictionary,
    author = {Hoàng Phê},
    title = {Từ điển tiếng Việt},
    publisher = {Nhà xuất bản Giáo dục},
    year = {2003}
}

@book{latin-viet-dictionary,
    author = {Alexandre de Rhodes},
    title = {Dictionarium Annamiticum Lusitanum et Latinum},
    year = {1651}
}

\appendix

\section{Appendix: Experimental Details}
\label{app:experimental-details}

\subsection{Metrics}
\label{app:metrics}

We evaluate model performance using complementary metrics that capture orthographic transcription accuracy, phonetic fidelity, dialectal robustness, and lexical generalization behavior.

\paragraph{Word Error Rate (WER).}
End-to-end transcription quality is primarily measured using \textit{Word Error Rate} (WER), defined as
\begin{equation}
\mathrm{WER} = \frac{S + D + I}{N},
\end{equation}
where $S$, $D$, and $I$ denote the numbers of word-level substitutions, deletions, and insertions obtained via minimum-edit-distance alignment, and $N$ is the number of words in the reference transcription. In addition to overall WER, we report dialect-wise WER for Northern, Central, and Southern Vietnamese, following the regional grouping defined in the UIT-ViMD benchmark.

\paragraph{Phone Error Rate (PER).}
For phonetic-level systems, we report \textit{Phone Error Rate} (PER), computed analogously at the phone level:
\begin{equation}
\mathrm{PER} = \frac{S_p + D_p + I_p}{N_p},
\end{equation}
where $S_p$, $D_p$, and $I_p$ are phone-level substitutions, deletions, and insertions, and $N_p$ is the number of reference phones. PER provides a linguistically grounded assessment of pronunciation modeling, which is particularly relevant for Vietnamese due to systematic dialectal variation at the phone level.

\paragraph{Component-wise Phonetic Error Rates.}
To analyze phonological modeling behavior with respect to Vietnamese syllable structure, we further compute component-wise error rates for syllable \textit{initials}, \textit{rhymes}, and \textit{tones}. For a component $c \in \{\text{initial}, \text{rhyme}, \text{tone}\}$, the error rate is defined as
\begin{equation}
\mathrm{ER}_c = \frac{S_c + D_c + I_c}{N_c},
\end{equation}
where $S_c$, $D_c$, $I_c$, and $N_c$ are computed over the aligned component sequence.

\paragraph{Dialect Classification Metrics.}
For multi-task models that jointly perform ASR and dialect identification, we report \textit{classification accuracy} and \textit{macro-averaged F1-score}. Given a set of dialect classes $C$, macro-F1 is computed as
\begin{equation}
\mathrm{Macro\text{-}F1} = \frac{1}{|C|} \sum_{c \in C} \mathrm{F1}_c,
\end{equation}
ensuring equal weighting across dialects regardless of class imbalance.

\paragraph{Lexical Diversity and Frequency Bias.}
To assess lexical generalization beyond aggregate error rates, we report the number of \textit{unique word types} correctly recognized at least once in the test set. We further analyze frequency bias by measuring the relationship between training-set word frequency and per-word recall.

For a word type $w$, its training-set frequency is defined as
\begin{equation}
f_{\text{train}}(w) =
\sum_{u \in \mathcal{D}_{\text{train}}}
\sum_{i=1}^{|u|}
\mathbf{1}\left(u_i = w\right),
\end{equation}
where $\mathcal{D}_{\text{train}}$ denotes the training corpus and $\mathbf{1}(\cdot)$ is the indicator function. To mitigate the heavy-tailed frequency distribution, we apply a logarithmic transformation:
\begin{equation}
\tilde{f}_{\text{train}}(w) = \log\left(1 + f_{\text{train}}(w)\right).
\end{equation}

Per-word recall is defined as
\begin{equation}
\mathrm{recall}(w) =
\frac{
\sum_{(x,y)\in \mathcal{D}_{\text{test}}}
\sum_{i=1}^{|\hat{y}|}
\mathbf{1}\left(\hat{y}_i = w\right)
}{
\sum_{(x,y)\in \mathcal{D}_{\text{test}}}
\sum_{i=1}^{|y|}
\mathbf{1}\left(y_i = w\right)
},
\end{equation}
where $(x,y)$ is a test utterance--reference pair and $\hat{y}$ is the predicted word sequence after optimal alignment.

The relationship between training frequency and recognition performance is quantified using Pearson and Spearman correlation coefficients:
\begin{equation}
r = \mathrm{Pearson}\left(\tilde{f}_{\text{train}}(w), \mathrm{recall}(w)\right),
\end{equation}
\begin{equation}
\rho = \mathrm{Spearman}\left(\tilde{f}_{\text{train}}(w), \mathrm{recall}(w)\right),
\end{equation}
computed over all word types appearing at least once in the test set. Lower correlation values indicate reduced dependence on frequency-driven memorization and stronger structural generalization.

\subsection{Additional Dialect Classification Results}

\begin{table*}[htp]
\centering
\small
\setlength{\tabcolsep}{6pt}
\begin{tabular}{c c c c @{\hspace{1cm}} c c c c}
\toprule
& \multicolumn{3}{c}{\textbf{Predicted Dialect}}
& & \multicolumn{3}{c}{\textbf{Predicted Dialect}} \\
\cmidrule(lr){2-4} \cmidrule(lr){6-8}
\textbf{True}
& \textbf{Northern} & \textbf{Central} & \textbf{Southern}
& \textbf{True}
& \textbf{Northern} & \textbf{Central} & \textbf{Southern} \\
\midrule
\multicolumn{4}{c}{\textbf{Conformer}}
& \multicolumn{4}{c}{\textbf{Transformer}} \\
\midrule
Northern & 711 & 0   & 13
& Northern & 714 & 0   & 10 \\
Central & 2   & 148 & 5
& Central & 0   & 152 & 3  \\
Southern & 58  & 3   & 847
& Southern & 57  & 6   & 845 \\
\bottomrule
\end{tabular}
\caption{Dialect confusion matrices for multi-task Transformer and Conformer models on the UIT-ViMD test set. Rows denote ground-truth dialects and columns denote predicted dialects.}
\label{tab:dialect_confusion}
\end{table*}

Table~\ref{tab:dialect_confusion} shows that the Central dialect achieves the highest classification accuracy among the three dialects for both Transformer and Conformer models, as reflected by the strong diagonal entries and minimal cross-dialect confusions. In contrast, most classification errors arise from confusion between the Northern and Southern dialects, which are mutually misclassified more frequently than either is with the Central dialect. This pattern is consistent across architectures and suggests that Central Vietnamese exhibits more distinctive acoustic-phonetic characteristics, while Northern and Southern dialects share partially overlapping phonological properties that make them harder to separate. Overall, the confusion matrices provide fine-grained evidence that complements the aggregate performance metrics in Table~\ref{tab:detect_eval}.

\section{Appendix: Phonetics and Orthography in Vietnamese} \label{app:phonetics-orthography}

Traditionally, the Vietnamese had the Nom alphabet as the main orthography system. However, the Nom alphabet was developed based on the ancient Han alphabet. This means we have to be fluent in the ancient Han to have the ability to use the Nom alphabet for reading and writing. Later on, \cite{latin-viet-dictionary} used the Latin alphabet to describe the speech sound of this language. This orthography system is simple and effective for describing almost all phonetic phenomena in Vietnamese. With its advantages that counterbalance its disadvantages, this Latin alphabet is gradually improved over time and finally becomes the national alphabet system of modern Vietnamese.

Although having the Nom alphabet or the Latin alphabet, these orthography systems all reflect two consistent characteristics of Vietnamese:

\begin{enumerate}
    \item Vietnamese is a monosyllabic language.
    \item The correspondence between graphemes and phonemes in Vietnamese is consistent.
\end{enumerate}
That is, in this language, we do not have the linking pronunciation as in English, and every phoneme has persistent writing forms. In Vietnamese, each syllable has three components: initials, rhymes, and tones. Rhyme has smaller components, which are glide, vowel, and final. We provide the list of all phonemes according to the syllabic structure of Vietnamese:

\begin{itemize}
    \item 22 phonemes as the initials:
    \begin{itemize}
        \item Plosive consonants: /\textipa{b, t, t\textsuperscript{h}}, k/.
        \item Fricative consonants: /\textipa{f, d, 7, z, j, s, \:s, \t{cC}, \t{t\:s}, \ng, x, v}/.
        \item Nasal consonants: /\textipa{n, m, \ng, \textltailn}/.
        \item Vibrant consonants: /\textipa{r, l}/.
    \end{itemize}
    
    \item 01 phonemes as the glide: /\textipa{\textsubarch{u}}/.
    
    \item 15 phonemes as the vowels: 
    \begin{itemize}
        \item Diphthongs: /\textipa{ie, uo, W9}/.
        \item Monophthongs: /\textipa{a, \u{a}, \u{9}, i, E, e, u, W, o, O, O:, 9}/.
    \end{itemize}
    
    \item 10 phonemes as the finals:
    \begin{itemize}
        \item Nasal consonants: /\textipa{n, t}/.
        \item Labial consonants: /\textipa{m, p}/.
        \item Velar consonants: /\textipa{\ng, k}/.
        \item Palatal consonants: /\textipa{\textltailn, c}/.
        \item Semivowels: /\textipa{\textsubarch{u}, \textsubarch{i}}/.
    \end{itemize}

    \item 6 phonemes denote tones:
    \begin{itemize}
        \item Flat tone is denoted by nothing.
        \item Low falling tone: /\textipa{\tone{22}\tone{11}}/.
        \item Mid raising tone: /\textipa{\tone{33}\tone{55}}/.
        \item Mid falling tone: /\textipa{\tone{33}\tone{11}}/.
        \item Mid glottalized-falling tone: /\textipa{\tone{33}P\tone{11}}/.
        \item Mid glottalized-raising tone: /\textipa{\tone{33}P\tone{55}}/.
    \end{itemize}
\end{itemize}

The writing form of these phonemes is consistent regardless of grammar. In particular:

\begin{itemize}
    \item 6 tones are denoted by a mark above or below the graphemes of vowels:
    \begin{itemize}
        \item Flat tone is denoted by nothing (a).
        \item Low falling tone /\textipa{\tone{22}\tone{11}}/ is denoted by a grave accent (à).
        \item Mid raising tone /\textipa{\tone{33}\tone{55}}/ is denoted by an acute accent (á).
        \item Mid falling tone /\textipa{\tone{33}\tone{11}}/ is denoted by a hook above (ả).
        \item Mid glottalized-falling tone /\textipa{\tone{33}P\tone{55}}/ is denoted by a tilde above (ã).
        \item Mid glottalized-raising tone /\textipa{\tone{33}P\tone{11}}/ is denoted by a dot below (ạ).
    \end{itemize}
    In the following texts, we provide writing forms of vowels regarding the mentioned phonemes. These examples might include the tone mark above or below the graphemes. Readers can discard these marks to see the true writing form of the vowels in Vietnamese. For instance, the dot below \textbf{iê} /\textipa{i9}/ in word \textbf{kiệm} /\textipa{kiem\tone{33}P\tone{55}}/ denotes the mid glottalized-raising tone /\textipa{\tone{33}P\tone{55}}/.
    
    \item 22 initials have 26 writing forms:
    \begin{itemize}
        \item b /\textipa{b}/. Eg: \textbf{b}a mẹ, \textbf{b}ánh kẹo, \textbf{b}uôn \textbf{b}án.
        \item t /\textipa{t}/. Eg: \textbf{t}âm \textbf{t}ư, \textbf{t}ịnh \textbf{t}iến, \textbf{t}ính cách.
        \item th /\textipa{t\textsuperscript{h}}/. Eg: \textbf{th}ách \textbf{th}ức, \textbf{th}ành \textbf{th}ạo.
        \item k, c, or q /\textipa{k}/. Eg: \textbf{c}ách mạng, \textbf{q}uan hệ, hiện \textbf{k}im.
        \item ph /\textipa{f}/. Eg: \textbf{ph}ụ huynh, \textbf{ph}ong cách, \textbf{ph}ân định.
        \item đ /\textipa{d}/. Eg: \textbf{đ}ưa \textbf{đ}ón, \textbf{đ}ậm \textbf{đ}à.
        \item gh or g /\textipa{7}/. Eg: \textbf{g}a tàu, \textbf{g}ánh hát, \textbf{g}anh \textbf{gh}ét.
        \item gi /\textipa{z}/. Eg: \textbf{gi}ếng nước, \textbf{gi}ống loài.
        \item đ /\textipa{d}/. Eg: \textbf{đ}ứng \textbf{đ}ắn, \textbf{đ}êm tối, \textbf{đ}èn \textbf{đ}uốc.
        \item d /\textipa{j}/. Eg: \textbf{d}a \textbf{d}ẻ, tiêu \textbf{d}ùng, \textbf{d}ụng cụ.
        \item x /\textipa{s}/. Eg: sản \textbf{x}uất, \textbf{x}uất \textbf{x}ứ, \textbf{x}e cộ.
        \item s /\textipa{\:s}/. Eg: xác \textbf{s}uất, \textbf{s}o \textbf{s}ánh, \textbf{s}ao chép.
        \item ch /\textipa{\t{cC}}/. Eg: \textbf{ch}ứa \textbf{ch}an, \textbf{ch}e \textbf{ch}ở/.
        \item tr /\textipa{\t{t\:s}}/. Eg: \textbf{tr}anh chấp, tiệt \textbf{tr}ùng.
        \item ng or ngh /\textipa{ng}/. Eg: mong \textbf{ng}óng, tình \textbf{ngh}ĩa.
        \item nh /\textipa{\textltailn}/. Eg: \textbf{nh}à cửa, nỗi \textbf{nh}ớ, \textbf{nh}ung lụa.
        \item l /\textipa{l}/. Eg: \textbf{l}ấm \textbf{l}em, \textbf{l}ung \textbf{l}inh, \textbf{l}ối về.
        \item r /\textipa{r}/. Eg: \textbf{r}ậm \textbf{r}ạp, \textbf{r}ón \textbf{r}én, \textbf{r}ực \textbf{r}ỡ.
        \item kh /\textipa{x}/. Eg: \textbf{kh}ó \textbf{kh}ăn, \textbf{kh}ởi sắc, \textbf{kh}ấm \textbf{kh}á.
        \item v /\textipa{v}/. Eg: \textbf{v}ui \textbf{v}ẻ, \textbf{v}ương \textbf{v}ấn, \textbf{v}ẫy \textbf{v}ùng.
        \item m /\textipa{m}/. Eg: \textbf{m}ong \textbf{m}ỏi, \textbf{m}ay \textbf{m}ắn, \textbf{m}ênh \textbf{m}ông.
        \item n /\textipa{n}/. Eg: đất \textbf{n}ước, \textbf{n}úi \textbf{n}on, \textbf{n}ông cạn.
    \end{itemize}

    \item The glide /\textipa{\textsubarch{u}}/ has two writing forms: u or o. Eg: q\textbf{u}ê nhà, h\textbf{o}a cỏ, kh\textbf{u}yến khích.

    \item 03 diphthongs have 8 writing forms:
    \begin{itemize}
        \item iê, yê, ia, or ya /\textipa{i@}/. Eg: kh\textbf{iế}m thị, \textbf{yê}n ắng, ch\textbf{ia} sẻ, khu\textbf{ya} khoắt.
        \item uô or ua /\textipa{uo}/. Eg: kh\textbf{uô}n khổ, m\textbf{ua} bán.
        \item ươ or ưa /\textipa{W@}/. Eg: kh\textbf{ướ}u giác, dây d\textbf{ưa}.
    \end{itemize}

    \item 12 monophthongs have 13 writing forms:
    \begin{itemize}
        \item a /\textipa{a}/. Eg: b\textbf{a} mẹ, tr\textbf{a}nh vẽ, ng\textbf{ã} b\textbf{a}, m\textbf{ải} miết, làng ch\textbf{ài}.
        \item ă or a /\textipa{ă}/. Eg: ánh \textbf{ắ}ng, n\textbf{ắ}m tay, n\textbf{ă}m tháng, \textbf{áy} n\textbf{áy}, chạy nhảy.
        \item â /\textipa{\u{@}}/. Eg: n\textbf{â}ng niu, \textbf{ấ}n định, ng\textbf{â}n vang.
        \item i or y /\textipa{i}/. Eg: th\textbf{i} cử, tr\textbf{ĩ}u nặng, b\textbf{ĩ}u môi.
        \item ê /\textbf{e}/. Eg: k\textbf{ế}t quả, thể hi\textbf{ệ}n, mân m\textbf{ê}.
        \item e /\textipa{E}/. Eg: mùa h\textbf{è}, x\textbf{e} cộ, t\textbf{é} ngã.
        \item u /\textipa{u}/. Eg: th\textbf{u} mua, m\textbf{ủ}m mỉm, l\textbf{u}ng lay, tr\textbf{u}ng thành.
        \item ư /\textipa{W}/. Eg: tr\textbf{ư}ng cầu, xây d\textbf{ự}ng, \textbf{ư}ng ý.
        \item o /\textipa{O}/. Eg: chăm s\textbf{ó}c, m\textbf{o}ng ng\textbf{ó}ng, tr\textbf{o}ng veo.
        \item oo /\textipa{O:}/. Eg: x\textbf{oo}ng chảo.
        \item ô /\textipa{o}/. Eg: tr\textbf{ố}ng đ\textbf{ồ}ng, \textbf{ố}ng hút, c\textbf{ố} nhân.
        \item ơ /\textipa{@}/. Eg: m\textbf{ơ} mộng, c\textbf{ơ} nh\textbf{ỡ}, ch\textbf{ơ}i b\textbf{ờ}i.
    \end{itemize}

    \item 10 final consonants have 12 writing forms:
    \begin{itemize}
        \item i or y /\textipa{\textsubarch{i}}/. Eg: làng chà\textbf{i}, mỏ\textbf{i} mệt, chạ\textbf{y} đua, bay nhả\textbf{y}.
        \item m /\textipa{m}/. Eg: ê\textbf{m} ấ\textbf{m}, nhiệ\textbf{m} màu, mâ\textbf{m} cỗ.
        \item n /\textipa{n}/. Eg: na\textbf{n} giải, no\textbf{n} nớt, tả\textbf{n} mạ\textbf{n}.
        \item ng /\textipa{\ng}/. Eg: sa\textbf{ng} trọ\textbf{ng}, trố\textbf{ng} trải, su\textbf{ng} túc.
        \item nh /\textipa{\textltailn}/. Eg: nha\textbf{nh} nhẹn, bê\textbf{nh} vực, bi\textbf{nh} quyền.
        \item p /\textipa{p}/. Eg: phậ\textbf{p} phồng, thấ\textbf{p} thỏm, thá\textbf{p} tùng, gượng é\textbf{p}, ức hiế\textbf{p}, tẩm ướ\textbf{p}.
        \item t /\textipa{t}/. Eg: lấn á\textbf{t}, bá\textbf{t} đĩa, kế\textbf{t} quả, hớ\textbf{t} hả.
        \item c /\textbf{k}/. Eg: cú\textbf{c} áo, chự\textbf{c} chờ, bố\textbf{c} vá\textbf{c}.
        \item ch /\textbf{c}/. Eg: cá\textbf{ch} thứ\textbf{c}, chí\textbf{ch} ngừa.
        \item u or o /\textipa{\textsubarch{u}}/. Eg: tra\textbf{u} chuốt, chị\textbf{u} đựng, xoong chả\textbf{o}, xiêu vẹ\textbf{o}.
    \end{itemize}
    
\end{itemize}

\begin{figure}
    \centering
    \includegraphics[width=\linewidth]{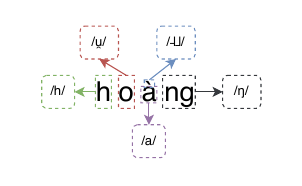}
    \caption{Example for the consistency between graphemes and phonemes in Vietnamese.}
    \label{fig:syllable-example}
\end{figure}

It is important to note that the writing form of phonemes in Vietnamese is consistent in all usage cases, regardless of grammar (tense, aspect, mood). This feature defines Vietnamese as an isolating language whose morphology is an aspect of phonetics rather than grammar, as in inflectional languages. To this end, given a Vietnamese word, native speakers can find no difficulty in pronouncing it. On the other hand, Vietnamese natives can write down any Vietnamese word by listening to its speech sound without knowing how to spell it. For instance, given the word \textbf{hoàng}, its phonetic representation is /\textipa{h\textsubarch{u}a\ng\tone{22}\tone{11}}/. We can determine the grapheme of the initial /\textipa{h}/ is \textbf{h}, of the glide /\textipa{\textsubarch{u}}/ is \textbf{o}, of the vowel /\textipa{a}/ is \textbf{a}, of the /\textipa{\ng}/ is \textbf{ng}, and tone is denoted by the grave accent above \textbf{à} (Figure \ref{fig:syllable-example}).

However, although the conversion from grapheme to phonemes is straightforward (that is, many graphemes correspond to unique phonemes), the inversion is not all ways many-to-one mapping. There are some phonemes having one-to-many mapping with graphemes, such as the initial /\textipa{7}/ can be written as \textbf{g} or \textbf{gh}, or the diphthong /\textipa{ie}/ can be written as \textbf{iê, yê, ia} or \textbf{ya}. Actually, a particular writing form of such phonemes is determined consistently via the neighbor phonemes. We detail here rules for determining writing forms of phonemes having multiple respective graphemes:

\begin{itemize}
    \item The initial /\textipa{k}/ is written as:
    \begin{itemize}
        \item \textbf{k} if followed by \textbf{i} /\textipa{i}/, \textbf{ê} /\textipa{e}/, \textbf{e} /\textipa{E}/, \textbf{iê} /\textipa{ie}/. Eg: \textbf{kị}p thời, \textbf{kiể}u cách, \textbf{kè}m cặp.
        \item \textbf{q} if followed by \textbf{u} /\textipa{\textsubarch{u}}/ as the glide. Eg: \textbf{qu}ê hương, \textbf{qu}à cáp.
        \item \textbf{c} otherwise. Eg: \textbf{cứ}ng \textbf{cỏ}i, \textbf{cở}i mở, hạt \textbf{cá}t, rau \textbf{củ}.
    \end{itemize}

    \item The initial /\textipa{7}/ is written as:
    \begin{itemize}
        \item \textbf{gh} if followed by \textbf{i} /\textipa{i}/, \textbf{ê} /\textipa{e}/, \textbf{e} /\textipa{E}/, \textbf{iê} /\textipa{ie}/. Eg: bàn \textbf{ghế}, \textbf{ghi} chép, \textbf{ghe} tàu.
        \item \textbf{g} otherwise. Eg: thanh gươm, gồng gánh, gọi điện.
    \end{itemize}

    \item The initial /\textipa{\ng}/ is written as:
    \begin{itemize}
        \item \textbf{ngh} if followed by textbf{i} /\textipa{i}/, \textbf{ê} /\textipa{e}/, \textbf{e} /\textipa{E}/, \textbf{iê} /\textipa{ie}/. Eg: \textbf{nghe} ngóng, \textbf{nghiê}m nghị, \textbf{nghệ} sĩ.
        \item \textbf{ng} otherwise. Eg: \textbf{ngà}nh nghề, \textbf{ngỗ} nghịch, \textbf{ngọ}t \textbf{ngà}o.
    \end{itemize}

    \item The diphthong /\textipa{ie}/ is written as:
    \begin{itemize}
        \item \textbf{iê} if the rhyme has a final consonant and no glide. Eg: k\textbf{iến} thức, tiết k\textbf{iệm}.
        \item \textbf{yê} if the rhyme has a final consonant and the glide written as \textbf{u}. Eg: kh\textbf{uyên} bảo, \textbf{uyển} ch\textbf{uyển}, câu ch\textbf{uyện}.
        \item \textbf{ya} if the rhyme has no final consonant and the glide written as \textbf{u}. Eg: đêm kh\textbf{uya}.
        \item \textbf{ia} if the rhyme has no final consonant and no glide. Eg: b\textbf{ìa} sách, ch\textbf{ia} sẻ.
    \end{itemize}

    \item The diphthong /\textipa{uo}/ is written as:
    \begin{itemize}
        \item \textbf{uô} if the rhyme has a final consonant. Eg: nỗi b\textbf{uồn}, m\textbf{uối} biển, m\textbf{uộn} màn, ch\textbf{uồn} ch\textbf{uồn}.
        \item \textbf{ua} if the rhyme has no final consonant. Eg: ch\textbf{ùa} chiền, nhảy m\textbf{úa}.
    \end{itemize}

    \item The diphthong /\textipa{W@}/ is written as:
    \begin{itemize}
        \item \textbf{ươ} if the rhyme has a final consonant. Eg: bia r\textbf{ượu}, h\textbf{ưởng} thụ, chiêm ng\textbf{ưỡng}.
        \item \textbf{ua} if the rhyme has no final consonant. Eg: ch\textbf{ùa} chiền, nhảy m\textbf{úa}.
    \end{itemize}

    \item The monophthong /\textipa{ă}/ is written as:
    \begin{itemize}
        \item \textbf{a} if is is followed by character \textbf{y}. Eg: m\textbf{áy} bay, c\textbf{ay} nồng, t\textbf{ay} chân.
        \item \textbf{ă} otherwise. Eg: b\textbf{ắt} tay, b\textbf{ằng} lòng, may m\textbf{ắn}.
    \end{itemize}

    \item The monophthong /\textipa{i}/ is written as:
    \begin{itemize}
        \item \textbf{i} if the rhyme has a final consonant. Eg: l\textbf{íu} r\textbf{ít}.
        \item \textbf{y} if the rhyme has no consonant. Eg: k\textbf{ỷ} luật, l\textbf{ý} do.
    \end{itemize}

    \item The final /\textipa{\textsubarch{i}}/ is written as:
    \begin{itemize}
        \item \textbf{i} if the rhyme has the front vowel /\textipa{a}/, the central vowel /\textipa{W, W9}/ or the back vowels /\textipa{u, o, O, uo}/ as the vowel. Eg: m\textbf{ải} mê, m\textbf{ui} thuyền, h\textbf{ỏi} han, m\textbf{ồi} chài, g\textbf{ửi} gắm, l\textbf{ười} biếng, n\textbf{uôi} nấng.
        \item \textbf{y} if the rhyme has /\textipa{\u{a}, \u{9}}/ as the vowel. Eg: b\textbf{ay} lượn, ch\textbf{ạy} nh\textbf{ảy}, c\textbf{ấy} c\textbf{ày}.
    \end{itemize}
\end{itemize}

Following these orthographic rules in Vietnamese, there is no ambiguity in converting phonemes to graphemes and vice versa. Analysis in this Section together with the analysis in the following Appendix \ref{app:multi-dialect} serve as the fundamentals for our Dialect-aware Tokenization algorithm, which is described comprehensively in Appendix \ref{app:tokenization}.

\section{Appendix: Multi-dialect Speech in Vietnamese}
\label{app:multi-dialect}

Most studies in multi-dialect speech in Vietnamese \cite{hao1998,thuat2016,chau-dialect} agree that Vietnamese has three dialects of speech, which are the Northern dialect, the Central dialect, and the Southern dialect. These dialects were studied and grouped mostly depending on how the residents deal with the rhymes. However, when diving into each particular dialect, there are more complicated phonetic phenomena, and more variations are explored. Section \ref{sec:preliminaries-examples} above briefly gives an overview of the main differences among the three main dialects in Vietnam. In this section, we provide the reader with a more detailed discussion and analysis of each dialect. These discussions form the fundamentals for our multi-dialect tokenization method, which is described in the following sections.

\subsection{Northern Dialect} \label{app:northern-dialect}

According to \cite{hao1998}, Northern dialect speech can be found in provinces from Thanh Hóa to the northern borders of Vietnam. In particular, the Northern dialect is distributed in Thanh Hóa, Hà Giang, Cao Bằng, Bắc Kạn, Lạng Sơn, Tuyên Quang, Thái Nguyên, Phú Thọ, Bắc Giang, Quảng Ninh, Lào Cai, Lai Châu, Yên Bái, Điện Biên, Sơn La, Hòa Bình, Hà Nội, Hà Nam, Bắc Ninh, Hải Dương, Hải Phòng, Hưng Yên, Nam Định, Ninh Bình, Thái Bình, and Vĩnh Phúc provinces.

In initials, Northern dialect does not give distinct pronunciation on the following phonemes \cite{hao1998,chau-dialect} (the writing forms are provided in the parentheses):

\begin{itemize}
    \item /\textipa{s}/ (x) and /\textipa{\:s}/ (s) are pronounced as [\textipa{s}].
    \item /\textipa{\t{cC}}/ (ch) and /\textipa{\t{t\:s}}/ (tr) are pronounced as [\textipa{\t{cC}}].
    \item /\textipa{z}/ (gi), /\textipa{r}/ (r), and /\textipa{j}/ (d) are pronounced as [\textipa{z}].
    \item In Hà Nội, Vĩnh Phúc, Bắc Ninh, Hà Đông, and Hưng Yên, there is a mispronunciation between /\textipa{l}/ (l) and /\textipa{n}/ (n).
\end{itemize}

In rhymes, this dialect has the general following features \cite{hao1998,chau-dialect}:

\begin{itemize}
    \item The glide /\textipa{\textsubarch{u}}/ does not follows /\textipa{u, o, O}/.
    \item The semivowel /\textipa{j}/ does not follow the front vowels /\textipa{i, e, E}/ except for the cases where the final consonants are velar.
    \item The back diphthongs /\textipa{W, W9}/ when being followed by the glide /\textipa{\textsubarch{u}}/ they are transmitted to the central diphthongs [\textipa{1, 19}]. That is the reason why the following words \textbf{con \textit{hươu}, ly \textit{rượu}, cấp \textit{cứu}, âm \textit{mưu}} are pronounced approximately the same as \textbf{con \textit{hiêu}, ly \textit{riệu}, cấp \textit{kíu}, âm \textit{miu}}, respectively.
\end{itemize}

\begin{figure}
    \centering
    \includegraphics[width=\linewidth]{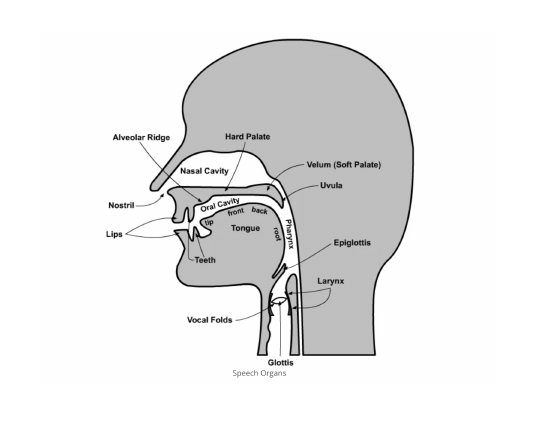}
    \caption{The speech organs.}
    \label{fig:speech-organs}
\end{figure}

In addition, this dialect exhibits two phonetic features on rhymes, including front vowels /\textipa{i, e, E}/ or rounded back vowels /\textipa{u, o, O}/ followed by velar consonants /\textipa{\ng, k}/. The front vowels require the tongue to be raised and forward to the teeth in the mouth, while the velar consonants require the tongue to be raised and backward to the velum. This results in the placement of the tongue in the middle of the mouth, not forward to the teeth or backward to the velum (Figure \ref{fig:speech-organs}). From that on, the velar consonants are palatalized to become [\textipa{\textltailn, c}] while the front vowels are transmitted to the central vowels [\textipa{1, 9, 3}]. The gliding between the central vowels and the palatalized consonants introduces a glide phone [\textipa{j}] in the middle. To this end, we have the following variations for the rhymes having front vowels /\textipa{i, e, E}/ followed by velar consonants /\textipa{\ng, k}/ (the writing forms are provided in the parentheses) \cite{hao1998,chau-dialect}:

\begin{itemize}
    \item /\textipa{i\ng}/ $\rightarrow$ [\textipa{1\textsuperscript{j}\textltailn}] (inh).
    \item /\textipa{e\ng}/ $\rightarrow$ [\textipa{9\textsuperscript{j}\textltailn}] (ênh).
    \item /\textipa{E\ng}/ $\rightarrow$ [\textipa{3\textsuperscript{j}\textltailn}] (anh).
    \item /\textipa{ik}/ $\rightarrow$ [\textipa{1\textsuperscript{j}c}] (ich).
    \item /\textipa{ek}/ $\rightarrow$ [\textipa{9\textsuperscript{j}c}] (êch).
    \item /\textipa{Ek}/ $\rightarrow$ [\textipa{3\textsuperscript{j}c}] (ach).
\end{itemize}

In the cases of the rounded back vowels /\textipa{u, o, O}/, the velar finals are labialized because of the effectiveness of labial vowels. These vowels are shortened according to the existence of the velar consonants, hence they lost the labial factor at the beginning. To this end, we have the following variations for the rhymes having rounded back vowels /\textipa{u, o, O}/ followed by velar consonants /\textipa{\ng, k}/ (the writing forms are provided in the parentheses) \cite{hao1998,chau-dialect}:

\begin{itemize}
    \item /\textipa{u\ng}/ $\rightarrow$ [\textipa{\textsuperscript{W}U\ng\textsuperscript{m}}] (ung).
    \item /\textipa{o\ng}/ $\rightarrow$ [\textipa{7\textsuperscript{w}\ng\textsuperscript{m}}] (ông).
    \item /\textipa{O\ng}/ $\rightarrow$ [\textipa{A\textsuperscript{w}\ng\textsuperscript{m}}] (ong).
    \item /\textipa{uk}/ $\rightarrow$ [\textipa{\textsuperscript{W}Uk\textsuperscript{p}}] (uc).
    \item /\textipa{ok}/ $\rightarrow$ [\textipa{7\textsuperscript{w}k\textsuperscript{p}}] (ôc).
    \item /\textipa{Ok}/ $\rightarrow$ [\textipa{A\textsuperscript{w}k\textsuperscript{p}}] (oc).
\end{itemize}

\subsection{Central Dialect} \label{app:central-dialect}

The Central dialect is distributed in more limited regions than the other two dialects. This dialect can be found in Nghệ An, Hà Tĩnh, Quảng Bình, Quảng Trị, and Huế provinces. Moreover, this dialect has a particular minor dialect for each province, which makes the Central dialect the most varied one in Vietnam. All minor variations of the Central dialect share the same features in initials and tones, while the rhymes are the only factor that makes them different from each other.

In general, this dialect has five tones:

\begin{itemize}
    \item In Quảng Bình, Quảng Trị, and Huế, there is no distinction between tone /\textipa{\tone{33}\tone{11}}/ and /\textipa{\tone{33}P\tone{55}}/.
    \item In Nghệ Tĩnh, there is no distinction between tone /\textipa{\tone{33}P\tone{55}}/ and /\textipa{\tone{33}P\tone{11}}/.
\end{itemize}
However, this dialect shares the same features relevant to initials in all provinces:

\begin{itemize}
    \item /\textipa{j, z}/ $\rightarrow$ [\textipa{z}].
    \item /\textipa{\t{t\:s}}/ $\rightarrow$ [\textipa{\:t}].
    \item /\textipa{\t{cC}}/ $\rightarrow$ [\textipa{c}].
\end{itemize}

The most significant factors that separate the minor dialects of the Central dialect are rhymes having front vowels /\textipa{i, e, E}/ followed by the velar consonants /\textipa{\ng, k}/. Rhymes having the rounded back vowels /\textipa{u, o, O}/ followed by the velar consonants in the Central dialect share the same color as those in the Northern dialect.

\subsubsection{Minor Dialect in Nghệ An and Hà Tĩnh} \label{app:nghe-an-ha-tinh}

Rhymes in Nghệ An and Hà Tĩnh reflect the same rules as the Northern dialect for those having front vowels followed by velar consonants. In these rhymes, the velar consonants /\textipa{\ng, k}/ are palatalized to become [\textipa{\textltailn, c}], but the front vowels keep their original form. To this end, we have:

\begin{itemize}
    \item /\textipa{i\ng}/ $\rightarrow$ [\textipa{i\textltailn}].
    \item /\textipa{e\ng}/ $\rightarrow$ [\textipa{e\textltailn}].
    \item /\textipa{E\ng}/ $\rightarrow$ [\textipa{E\textltailn}].
    \item /\textipa{ik}/ $\rightarrow$ [\textipa{ic}].
    \item /\textipa{ek}/ $\rightarrow$ [\textipa{ec}].
    \item /\textipa{Ek}/ $\rightarrow$ [\textipa{Ec}].
\end{itemize}

\subsubsection{Minor Dialect in Quảng Trị}

However, the rhymes mentioned in the above Section \ref{app:nghe-an-ha-tinh} are pronounced slightly differently from those in Quảng Trị. In this province, both front vowels and velar consonants keep their original form:

\begin{itemize}
    \item /\textipa{i\ng}/ $\rightarrow$ [\textipa{i\ng}].
    \item /\textipa{e\ng}/ $\rightarrow$ [\textipa{e\ng}].
    \item /\textipa{E\ng}/ $\rightarrow$ [\textipa{E\ng}].
    \item /\textipa{ik}/ $\rightarrow$ [\textipa{ik}].
    \item /\textipa{ek}/ $\rightarrow$ [\textipa{ek}].
    \item /\textipa{Ek}/ $\rightarrow$ [\textipa{Ek}].
\end{itemize}

\subsubsection{Minor Dialect in Quảng Bình}

In Quảng Bình, for rhymes having the nasal consonants /\textipa{n, t}/ following the front vowels /\textipa{i, e, E}/, these vowels are pronounced longer than usual:

\begin{itemize}
    \item /\textipa{in}/ $\rightarrow$ [\textipa{i:n}].
    \item /\textipa{en}/ $\rightarrow$ [\textipa{e:n}].
    \item /\textipa{En}/ $\rightarrow$ [\textipa{E:n}].
    \item /\textipa{it}/ $\rightarrow$ [\textipa{i:t}].
    \item /\textipa{et}/ $\rightarrow$ [\textipa{e:t}].
    \item /\textipa{Et}/ $\rightarrow$ [\textipa{E:t}].
\end{itemize}

However, these vowels (\textipa{i:, e:, E:}) become shorter (\textipa{i, e, E}) when being followed by the velar consonants /\textipa{\ng, k}/. From that on, the distinction between these pairs /\textipa{in}/ - /\textipa{i\ng}/, /\textipa{en}/ - /\textipa{e\ng}/, /\textipa{En}/ - /\textipa{E\ng}/, /\textipa{it}/ - /\textipa{ik}/, /\textipa{et}/ - /\textipa{ek}/, /\textipa{Et}/ - /\textipa{Ek}/ is not necessarily dependent on the final but on the length of the vowel. To this end, this dialect replaces the velar consonants /\textipa{\ng, k}/ by the nasal consonants /\textipa{n, t}/. These result in the following variations:

\begin{itemize}
    \item /\textipa{i\ng}/ $\rightarrow$ [\textipa{in}].
    \item /\textipa{e\ng}/ $\rightarrow$ [\textipa{en}].
    \item /\textipa{E\ng}/ $\rightarrow$ [\textipa{En}].
    \item /\textipa{ik}/ $\rightarrow$ [\textipa{it}].
    \item /\textipa{ek}/ $\rightarrow$ [\textipa{et}].
    \item /\textipa{Ek}/ $\rightarrow$ [\textipa{Et}].
\end{itemize}

\subsubsection{Minor Dialect in Huế}

In this minor dialect, for rhymes having front vowels /\textipa{i, e, E}/ followed by velar consonants /\textipa{\ng, k}/, these vowels are transmitted to the central vowels [\textipa{1, 9, 3}]. However, as the distinction between these pairs /\textipa{i\ng}/ - /\textipa{in}/, /\textipa{ik} - \textipa{it}/, /\textipa{e\ng} - \textipa{en}/, /\textipa{ek} - \textipa{et}/, /\textipa{E\ng} - \textipa{En}/, /\textipa{Ek} - \textipa{Et}/ largely depends on the difference of the vowels, hence having the velar consonants for these rhymes /\textipa{i\ng, ik, e\ng, ek, E\ng, Ek}/ is not necessary. To this end, these velar consonants are replaced by the respective nasal consonants /\textipa{n, t}/. From then on, we have:

\begin{itemize}
    \item /\textipa{i\ng}/ $\rightarrow$ [\textipa{1n}].
    \item /\textipa{ik}/ $\rightarrow$ [\textipa{1t}].
    \item /\textipa{e\ng}/ $\rightarrow$ [\textipa{9n}].
    \item /\textipa{ek}/ $\rightarrow$ [\textipa{9t}].
    \item /\textipa{E\ng}/ $\rightarrow$ [\textipa{3n}].
    \item /\textipa{Ek}/ $\rightarrow$ [\textipa{3t}].
\end{itemize}
The front vowels /\textipa{i, e, E}/ preceding the nasal consonants /\textipa{n, t}/become longer. That is:

\begin{itemize}
    \item /\textipa{in}/ $\rightarrow$ [\textipa{i:n}].
    \item /\textipa{it}/ $\rightarrow$ [\textipa{i:t}].
    \item /\textipa{en}/ $\rightarrow$ [\textipa{e:n}].
    \item /\textipa{et}/ $\rightarrow$ [\textipa{e:t}].
    \item /\textipa{En}/ $\rightarrow$ [\textipa{E:n}].
    \item /\textipa{Et}/ $\rightarrow$ [\textipa{E:t}].
\end{itemize}

From these analyses, we can see that the Huế dialect exhibits the same behavior as the dialect in Quảng Bình. However, the way of pronouncing rhymes having the front vowels followed by the velar consonants is different in these two minor dialects. These contribute to the huge diversity of dialects in Vietnamese, which results in a significant challenge in modeling multi-dialect speech ASR in this language.

\subsection{Southern Dialect} \label{app:southern-dialect}

\begin{figure}[t]
    \centering
    \includegraphics[width=\linewidth]{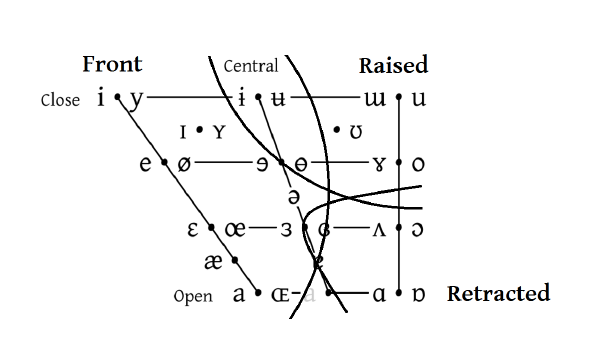}
    \caption{Vowel diagram.}
    \label{fig:vowel-diagram}
\end{figure}

\cite{hao1998,chau-dialect} determined that the southern dialect can be found from the Đà Nẵng province to the southern border of Vietnam. In particular, the Southern dialect is distributed in Đà Nẵng, Quảng Nam, Quảng Ngãi, Kom Tum, Bình Định, Gia Lai, Phú Yên, Đắk Lắk, Khánh Hòa, Lâm Đồng, Ninh Thuận, Bình Thuận, Bình Phước, Đồng Nai, Bình Dương, Bà Rịa - Vũng Tàu, Hồ Chí Minh, Long An, Đồng Tháp, Tiền Giang, Bến Tre, Vĩnh Long, An Giang, Trà Vinh, Cần Thơ, Sóc Trăng, Kiên Giang, Bạc Liêu, and Cà Mau.

This dialect has five tones rather than six tones as the Northern dialect: there is no distinction between tone /\textipa{\tone{33}\tone{55}}/ and /\textipa{\tone{33}P\tone{55}}/. In initials, Southern dialect does not give any difference between the following phonemes \cite{chau-dialect,hao1998}:

\begin{itemize}
    \item /\textipa{z}/ (gi), /\textipa{j}/ (d), and /\textipa{v}/ are all pronounced as [\textipa{j}].
    \item /\textipa{k\textsuperscript{w}}/ (qu) is pronounced as [\textipa{w}].
    \item This dialect also exhibits the same phonetic rule as the Northern dialect, where the velar consonants /\textipa{\ng, k}/ follow the front vowels /\textipa{i, e, E}/. These velar consonants are palatalized to become [\textipa{\textltailn}, c], but the front vowels /\textipa{i, e, E}/ keep their original.
    \item In provinces Quảng Nam, Quảng Ngãi, Kon Tum, Gia Lai, Đắk Lắk, Bình Định, Phú Yên, Khánh Hòa, Ninh Thuận, Bình Thuận, and Lâm Đồng: /\textipa{\t{t{\:s}}}/ (tr) is pronounced as [\textipa{\:t}]. While in other provinces, /\textipa{\t{t{\:s}}}/ (tr) and /\textipa{\t{cC}}/ (ch) is pronounced as [\textipa{c}], /\textipa{\:s}/ (s) and /\textipa{s}/ (x) are all pronounced as [\textipa{s}].
\end{itemize}

In rhymes, when these diphthongs /\textipa{ie, W9}/ followed by the semivowel /\textipa{\textsubarch{i}}/ or labial consonants /\textipa{m, p, \textsubarch{u}}/, they are converted to the monophthongs in the same row of the vowel diagram (Figure \ref{fig:vowel-diagram}). In particular, in this dialect we have:

\begin{itemize}
    \item /\textipa{iep}/ $\rightarrow$ [\textipa{ip}].
    \item /\textipa{W9p}/ $\rightarrow$ [\textipa{Wp}].
    \item /\textipa{iem}/ $\rightarrow$ [\textipa{im}].
    \item /\textipa{W9m}/ $\rightarrow$ [\textipa{Wm}].
    \item /\textipa{ie\textsubarch{u}}/ $\rightarrow$ [\textipa{i\textsubarch{u}}].
    \item /\textipa{W9\textsubarch{u}}/ $\rightarrow$ [\textipa{W\textsubarch{u}}].
\end{itemize}
This is the reason why in Southern dialect, the following words \textbf{\textit{tiếp} tục, quả \textit{mướp}, đánh \textit{chiếm}, \textit{ươm} mầm, tình \textit{yêu}, con \textit{hươu}} are approximately pronounced as \textbf{\textit{típ} tục, quả \textit{mứp}, đánh \textit{chím}, \textit{ưm} mầm, tình \textit{iu}, con \textit{hưu}}, respectively.

In addition to the general features of Southern dialect, two minor dialects share the same characteristics of Southern dialect but have their own particular features. The first one is the dialect distributed in provinces Quảng Nam and Quảng Ngãi, and the second one is the dialect found in the Mekong Delta.

\subsubsection{Minor Dialect of Southern Dialect in Quảng Nam and Quảng Ngãi}

In provinces Quảng Nam and Quảng Ngãi, the residents additionally pronounce the vowels differently from the Southern dialect. In rhymes without final consonants, the close monophthongs /\textipa{i, u, W}/ are converted into diphthongs whose the first components are the respective more-open vowels (the row below of the vowel diagram, Figure \ref{fig:vowel-diagram}), while the second components are the respective glides. In particular, we have:

\begin{itemize}
    \item /\textipa{i}/ $\rightarrow$ [\textipa{Ij}].
    \item /\textipa{u}/ $\rightarrow$ [\textipa{Uw}].
    \item /\textipa{W}/ $\rightarrow$ [\textipa{7W}].
\end{itemize}

In rhymes having /\textipa{\textsubarch{i}}/ or /\textipa{\textsubarch{u}}/ as the final consonants, the diphthongs /\textipa{uo, W9, ie}/ are transmitted into the monophthongs on the same row of the vowel diagram, but these monophthongs are longer to match the length of the original diphthongs:

\begin{itemize}
    \item /\textipa{uo}/ $\rightarrow$ [\textipa{u:}].
    \item /\textipa{W@}/ $\rightarrow$ [\textipa{W:}].
    \item /\textipa{ie}/ $\rightarrow$ [\textipa{i:}].
\end{itemize}
Moreover, the following two rhymes are pronounced totally differently:

\begin{itemize}
    \item /\textipa{oi}/ $\rightarrow$ [\textipa{u9}].
    \item /\textipa{ai}/ $\rightarrow$ [\textipa{ae}].
\end{itemize}
This is the reason why people in these two provinces pronounce the following words \textbf{\textit{nói}} and \textbf{\textit{hai}} approximately the same as \textbf{\textit{núa}} and \textbf{\textit{he}}, respectively.

Furthermore, in rhymes having labial finals /\textipa{m, p}/ and the rounded semivowel /\textipa{\textsubarch{u}}/, the vowels in this dialect are pronounced as follows:

\begin{itemize}
    \item /\textipa{a}/ $\rightarrow$ [\textipa{O}].
    \item /\textipa{O}/ $\rightarrow$ [\textipa{o}].
    \item /\textipa{o}/ $\rightarrow$ [\textipa{7}].
    \item /\textipa{u}/ $\rightarrow$ [\textipa{U}].
    \item /\textipa{\u{a}, \u{9}}/ $\rightarrow$ [\textipa{a}].
    \item In rhymes having /\textipa{m}/ as the final, /\textipa{uo}/ $\rightarrow$ [\textipa{U}].
\end{itemize}

In rhymes having the velar consonants as the final, while the front vowels /\textipa{i, e, E}/ do not vary, the following vowel changed as follows:

\begin{itemize}
    \item /\textipa{a}/ $\rightarrow$ [\textipa{A:}].
    \item /\textipa{O}/ $\rightarrow$ [\textipa{A:\textsuperscript{m}}].
    \item /\textipa{o}/ $\rightarrow$ [\textipa{2}].
    \item /\textipa{\u{@}}/ $\rightarrow$ [\textipa{A}].
    \item /\textipa{uo}/ $\rightarrow$ [\textipa{u:}].
    \item /\textipa{ie}/ $\rightarrow$ [\textipa{i:}].
    \item /\textipa{W@}/ $\rightarrow$ [\textipa{W:}].
\end{itemize}

\subsubsection{Minor Dialect of Southern Dialect in Mekong Delta}

Mekong Delta includes Bình Dương, Bình Phước, Đồng Nai, Bà Rịa - Vũng Tàu, Tây Ninh, Hồ Chí Minh, Long An, Tiền Giang, Hậu Giang, Long An, Bến Tre, Đồng Tháp, Vĩnh Long, Trà Vinh, Cần Thơ, Sóc Trăng, An Giang, Kiên Giang, Bạc Liêu, and Cà Mau. These provinces have a unique dialect that shares many features with the Southern dialect, with their own distinctions.

In particular, the front vowels /\textipa{i, e}/ followed by velar consonants /\textipa{\ng, k}/ do not transmitted to the respective central vowels, but the velar consonants are replaced by the nasal consonants /\textipa{\ng, k}/ $\rightarrow$ /\textipa{n, t}/, which is similar to the Central dialect without changing in length of the front vowels. That is:

\begin{itemize}
    \item /\textipa{i\ng}/ $\rightarrow$ [\textipa{in}].
    \item /\textipa{ik}/ $\rightarrow$ [\textipa{it}].
\end{itemize}
In addition, the front vowel /\textipa{e}/ followed by the velar consonants /\textipa{\ng, k}/ is converted to the respective central vowel because of the velar consonants, but as the velar consonants /\textipa{\ng, k}/ are replaced by the nasal consonants [\textipa{n, t}], which results in this vowel is longer. From then on, we have:

\begin{itemize}
    \item /\textipa{e\ng}/ $\rightarrow$ [\textipa{9:n}].
    \item /\textipa{ek}/ $\rightarrow$ [\textipa{9:t}].
\end{itemize}

However, the alveolar vowels /\textipa{n, t}/ following vowel /\textipa{a}/, and the central vowels /\textipa{9, E}/ become the velar consonants /\textipa{\ng, k}/. This is the reason why in this dialect, these words \textbf{buôn \textit{bán}}, \textbf{\textit{ít} ỏi}, \textbf{\textit{ân cần}}, \textbf{\textit{chen lấn}} are pronounced approximately the same as \textbf{buôn \textit{báng}}, \textbf{\textit{ích} ỏi}, \textbf{\textit{âng cầng}}, \textbf{\textit{cheng lấng}}, respectively.

\begin{figure*}[htp]
    \centering
    \includegraphics[width=0.65\linewidth]{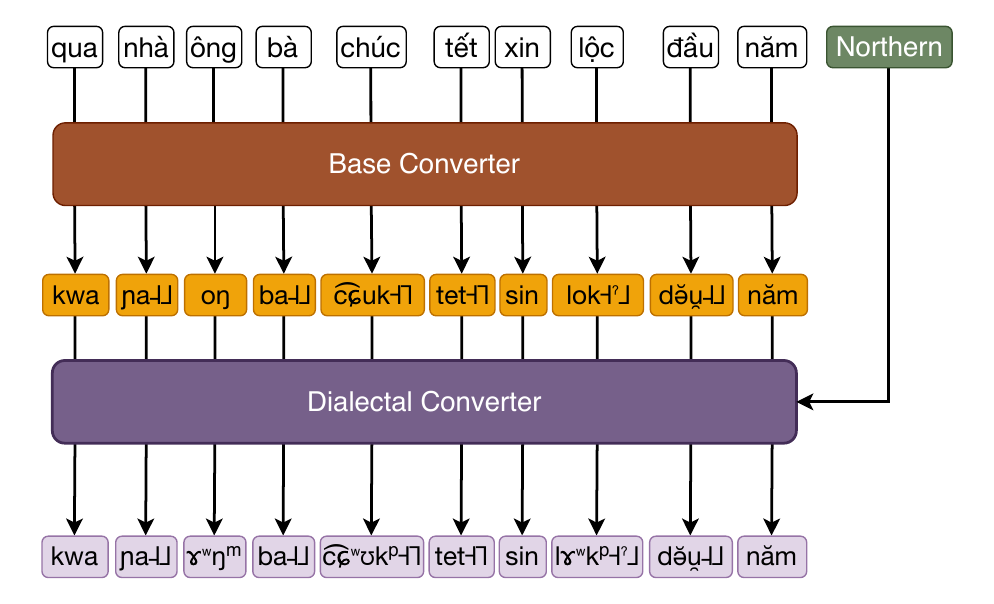}
    \caption{Dialect-aware tokenization algorithm.}
    \label{fig:tokenization}
\end{figure*}

On the other hand, the rounded back ones /\textipa{u, o, O)}/ have variations depending on the following consonants. If these vowels are followed by the nasal consonants /\textipa{n, t}/, they keep their rounded features while their length is longer /\textipa{u, o, O}/ $\rightarrow$ [\textipa{u:, o:, O:}]. In the case these vowels are followed by velar consonants /\textipa{\ng, k}/, they lose the rounded characteristics as in the Northern dialect with a little bit difference, while these velar consonants are labialized to become [\textipa{\ng\textsuperscript{m}, k\textsuperscript{p}}]. Moreover, the nasal consonants [\textipa{n, t}] followed by these vowels are replaced by the velar-labialized consonants [\textipa{\ng\textsuperscript{m}, k\textsuperscript{p}}], respectively. In particular, we have:

\begin{itemize}
    \item /\textipa{un, u\ng}/ $\rightarrow$ [\textipa{\textsuperscript{W}U\ng\textsuperscript{m}}].
    \item /\textipa{ut, uk}/ $\rightarrow$ [\textipa{\textsuperscript{W}Uk\textsuperscript{p}}].
    \item /\textipa{On, O\ng}/ $\rightarrow$ [\textipa{7\ng\textsuperscript{m}}].
    \item /\textipa{Ot, Ok}/ $\rightarrow$ [\textipa{2k\textsuperscript{p}}].
\end{itemize}

\section{Appendix: Additional Results on Subword-Level ASR Models}

\begin{table}[ht]
    \centering
    \resizebox{0.45\textwidth}{!}{
    \begin{tabular}{lccc}
    \hline
    \textbf{Model} & \textbf{Setup} & \textbf{WER} & \textbf{PER} \\ \hline
    \multirow{5}{*}{Transformer} & word-level & 16.71 $\pm$ 0.13 & - \\
     & BPE $\sim$273 & 17.16 $\pm$ 0.11 & - \\
     & BPE $\sim$14k & 15.31 $\pm$ 0.06 & - \\
     & P & 15.53 $\pm$ 0.08 & 12.47 $\pm$ 0.16 \\
     & P+V & 13.05 $\pm$ 0.09 & 8.36 $\pm$ 0.12 \\ \hline
    \multirow{5}{*}{Conformer} & word-level & 17.49 $\pm$ 0.08 & - \\
     & BPE $\sim$273 & 17.53 $\pm$ 0.13 & - \\
     & BPE $\sim$14k & 15.77 $\pm$ 0.08 & - \\
     & P & 13.89 $\pm$ 0.11 & 11.22 $\pm$ 0.09 \\
     & P+V & 13.27 $\pm$ 0.16 & 8.47 $\pm$ 0.14 \\ \hline
    \end{tabular}}
    \caption{Experimental results of Transformer and Conformer using Subword-level BPE tokenizer and our Dialect-aware Phonetic Representation. \textbf{BPE$\sim$273} denotes methods having BPE vocabs of 273 tokens, while \textbf{BPE$\sim$14k} denotes methods having BPE vocabs of 14,670 tokens. \textbf{P} denotes phonetic supervision and \textbf{V} denotes methods having Phonetic-Structure Decoder.}
    \label{tab:bpe-results}
\end{table}

To evaluate the effect of our phonetic representation on the performance of Transformer and Conformer models compared with the subword-based BPE tokenizer, we additionally trained these two architectures using BPE vocabularies of two different sizes: 273 tokens and 14,670 tokens. The reason for selecting this vocabulary size is that 273 is the vocab size of our Dialect-aware Phonetic Vocabulary, and 14,670 is the maximal size that BPE algorithm of SentencePiece\footnote{https://github.com/google/sentencepiece} suggests to construct the vocab on the UIT-ViMD dataset. The Transformer and Conformer configurations for both BPE settings were kept identical to those used in the main experiments.

As shown in Table~\ref{tab:bpe-results}, the Transformer model with BPE$\sim$273 achieved lower performance than the word-level counterpart. Similarly, the Conformer model with BPE$\sim$273 produced a higher WER than the Conformer using word-level tokens. This degradation can be attributed to the limited vocabulary size, which prevents BPE from representing complete words and instead forces it to decompose words into sequences of characters. Our analysis of the BPE tokenizer with a vocabulary size of 273 revealed that most generated subwords are not the completed words in Vietnamese, resulting in an inadequate representation for Vietnamese text.

In contrast, Transformer and Conformer models trained with BPE$\sim$14k achieved substantially lower WERs than their word-level counterparts. In particular, the Transformer using BPE$\sim$14k outperformed the word-level Transformer by approximately 1.4\% absolute WER. A similar trend was observed for the Conformer architecture, where the BPE$\sim$14k model achieved a WER of 15.77\%, compared to 17.49\% for the word-level Conformer.

Despite the considerable improvements obtained with the large-vocabulary BPE tokenizer, the Transformer and Conformer models using our phonetic vocabulary still consistently outperformed both the BPE$\sim$14k and word-level systems by significant margins. These results further demonstrate the effectiveness of our proposed phonetic representation for modeling Vietnamese dialectal speech, compared with conventional orthographic representations.

\section{Appendix: Reverse-Lexicon Analysis} \label{app:reverse}

We indicated in Appendix \ref{app:phonetics-orthography}), the conversion from phonemes back to graphemes is one-to-one mapping thanks to the high corresponding between phonetics and orthography of Vietnamese. However, the conversion between phones to phonemes is ambiguity across dialects. For instance, in Northern dialect (Appendix \ref{app:northern-dialect}, both /\textipa{s}/ and /\textipa{\:s}/ are pronounced as [\textipa{s}]. This introduces an ambiguity of converting [\textipa{s}] back to the correct phonemes. Let us analyze every cases so that we can conclude the how much the ambiguity of the reverse-lexicon process.

\subsection{Reverse-Lexicon Analysis of Northern Dialect}

There exist several initials in this dialect that are realized as the same phone. For example, given the phonetic sequence [\textipa{sa\ng}], the corresponding phonemic forms can be either /\textipa{sa\ng}/ or /\textipa{\:sa\ng}/, which map to the orthographic forms \textit{sang} and \textit{xang}, respectively. However, \textit{xang} is not a valid Vietnamese word, whereas \textit{sang} is a valid lexical item. 

To resolve such ambiguities, we propose a two-step approach for recovering the correct orthographic form from the phonetic representation: (1) generating all possible phonemic forms together with their corresponding orthographic forms, and (2) leveraging a standard Vietnamese dictionary \cite{phe-dictionary} to identify the valid orthographic candidate. In practice, we found that most ambiguous cases involving initials in this dialect can be effectively resolved using the standard Vietnamese dictionary, except for:

\begin{itemize}
    \item 18 triplets of words, approximately 0.9\% of total single words in standard Vietnamese dictionary, starting with /\textipa{z}/ - /\textipa{j}/ - /\textipa{r}/.
    \item 26 pairs of words, approximately 0.8\% of total single words in standard Vietnamese dictionary, starting with /\textipa{\t{cC}}/ - /\textipa{\t{t\:s}}/.
    \item 19 pairs of words, approximately 0.6\% of total single words in standard Vietnamese dictionary, starting with /\textipa{s}/ - /\textipa{\:s}/.
\end{itemize}

For vowel restoration, this dialect mainly exhibits variations involving front vowels followed by velar consonants and back vowels followed by velar consonants. As shown in Appendix~\ref{app:northern-dialect}, the mapping between phones and phonemes is bijective in these cases. Consequently, recovering phonemes and subsequently graphemes from vowels is straightforward and does not introduce ambiguity. The same findings applies for the cases of vowels having diphthongs.

\subsection{Reverse-Lexicon Analysis of Central Dialect}

Unlike the Northern dialect, the Central dialect contains five tones instead of six. As a result, multiple words may share the same pronunciation due to tonal mergers. In addition, ambiguities may also arise from the realization of initials (Appendix~\ref{app:central-dialect}). Nevertheless, the ambiguities can be comprehensively resolved by leveraging a standard Vietnamese dictionary to determine the valid phonemic forms corresponding to the tones in this dialect. For the cases of initials sharing the same pronunciation, our statistics show that there are

\begin{itemize}
    \item 19 pairs of words, approximately 0.6\% of total single words in standard Vietnamese dictionary, starting with /\textipa{s}/ - /\textipa{\:s}/.
    \item 10 pairs of words, approximately 0.2\% of total single words in standard Vietnamese dictionary, starting with /\textipa{z}/ - /\textipa{j}/.
\end{itemize}

The remaining ambiguous cases involve the front vowels /\textipa{i, e, E}/ and back vowels /\textipa{u, o, O}/ followed by the velar consonants /\textipa{\ng, k}/. In the Quảng Trị minor-dialect, these vowels preserve their original forms. In the Nghệ An and Hà Tĩnh subdialects, only the finals are modified according to the preceding phonemes; however, the mapping between phones and phonemes remains bijective. Similarly, the vowel variations observed in the Huế and Quảng Bình subdialects also yield bijective mappings. Therefore, converting vowels back to phonemes in the Central dialect does not introduce linguistic ambiguity.

\subsection{Reverse-Lexicon Analysis of Southern Dialect}

Similar to Central dialect, Southern dialect remains five tones instead of six. There are also some initials having the same pronunciation such as /\textipa{s}/ and /\textipa{\:s}/, or /\textipa{z}/, /\textipa{j}/, and /\textipa{v}/. For instance, two syllables /\textipa{jW@\tone{22}\tone{11}}/ (\textit{dừa}) and /\textipa{vW@\tone{22}\tone{11}}/ (\textit{vừa}) share the same phonetic form [\textipa{jW@\tone{22}\tone{11}}]. Fortunately, the tonal confusion can be solved comprehensively using the standard Vietnamese dictionary as in the Northern dialect, while the ambiguity of the initials remains.

Our analysis shows there are 34 triplets of words starting with /\textipa{v}/ - /\textipa{j}/ - /\textipa{z}/ that sharing the same phonetic initial /\textipa{j}/ (Appendix \ref{app:southern-dialect}), where:

\begin{itemize}
    \item 13 pairs of words starting with /\textipa{v}/ - /\textipa{z}/.
    \item 31 pairs of words starting with /\textipa{v}/ - /\textipa{j}/.
    \item 10 pairs of words starting with /\textipa{j}/ - /\textipa{z}/.
\end{itemize}
and these words occupy \textbf{approximately 0.8\%} of the standard Vietnamese dictionary. For the confusion of the initial [\textipa{s}], there are 19 pairs of words that start with /\textipa{s}/ - /\textipa{\:s}/, and they occupy \textbf{approximately 0.3\%} of the standard Vietnamese dictionary. Finally, for the confusion of the initial /\textipa{c}/ in some provinces, there are 26 pairs of words starting with /\textipa{\t{t{\:s}}}/ - /\textipa{\t{cC}}/ which occupy 0.4\%.

Relevant to the confusion of vowels, according to Appendix \ref{app:southern-dialect}, we have:

\begin{itemize}
    \item /\textipa{in, i\ng}/ $\rightarrow$ [\textipa{in}].
    \item /\textipa{ik, it}/ $\rightarrow$ [\textipa{it}].
    \item /\textipa{un, u\ng}/ $\rightarrow$ [\textipa{\textsuperscript{W}U\ng\textsuperscript{m}}].
    \item /\textipa{ut, uk}/ $\rightarrow$ [\textipa{\textsuperscript{W}Uk\textsuperscript{p}}].
    \item /\textipa{On, O\ng}/ $\rightarrow$ [\textipa{7\ng\textsuperscript{m}}].
    \item /\textipa{Ot, Ok}/ $\rightarrow$ [\textipa{2k\textsuperscript{p}}].
\end{itemize}
In Vietnamese, there are approximately 10.59\% of standard words having these vowels in their syllabic structure.

\subsection{Reverse-Lexicon Analysis in Conclusion}

We further conducted a detailed analysis of the ambiguities introduced during the conversion process from phones to phonemes and subsequently to graphemes in our tokenizer. The analysis shows that, for the Northern dialect, approximately 2.23\% of words may lead to ambiguity, implying that around half of these cases, i.e., \textbf{1.115\%}, are incorrectly reconstructed. 

For the Central dialect, only 0.8\% of words introduce ambiguity, resulting in approximately \textbf{0.4\%} of words being converted into incorrect orthographic forms. In contrast, the Southern dialect exhibits a higher ambiguity rate, where approximately 12.09\% of words may cause confusion, implying that around \textbf{6.045\%} are decoded incorrectly.

Overall, approximately \textbf{7.56\%} of words are incorrectly reconstructed when converting from phonetic representations back to orthographic forms using our tokenizer. Although this error rate remains relatively small, it is still an important limitation that should be addressed and further improved in future studies, particularly for Vietnamese speech processing.

\section{Appendix: Dialect-aware Phonetic Vocabulary} \label{app:tokenization}

We have analyzed in detail the phonetic and orthographic features of Vietnamese in Section \ref{app:phonetics-orthography} and comprehensively describe the multi-dialect variations of this language in Section \ref{app:multi-dialect}. From these analysis, we develop the Vietnamese Dialect Tokenization (ViDialectToken) algorithm. ViDialectToken receives a line of text which is the transcription of the given multi-dialect voice and the province ID, then processes and returns the sequence of IPA (International Phonetic Alphabet) characters describing the dialect speech sound of the given audio correspond to the dialect. ViDialectToken has two phases (Figure \ref{fig:tokenization}): (1) Retrieving phonemes from graphemes given the text of the transcript (the Base Converter module) and (2) Converting phonemes to IPA characters respective to dialectal phones (the Dialectal Converter module).

\subsection{Base Converter Module}

\SetKwFunction{FTone}{get\_tone}
\SetKwFunction{FInitial}{get\_initial}
\SetKwFunction{FGlide}{get\_glide}
\SetKwFunction{FVowel}{get\_vowel}
\SetKwFunction{FFinal}{get\_final}

\begin{algorithm}[t]
\caption{The algorithm for converting text to phonemes.}\label{alg:text2phoneme}
\LinesNumbered
\KwData{Transcript of the audio $w = (w_1, w_2, ..., w_n)$.}
\KwResult{A sequence of syllables $p = (p_1, p_2, ..., p_n)$ of the given input transcript $w = (w_1, w_2, ..., w_n)$. Each phoneme $p_i = (p_i^{init}, p_i^{glide}, p_i^{vowel}, p_i^{final}, p_i^{tone})$ is a triplet of IPA for the initial, rhyme, and tone.}

phonemes $\leftarrow$ an empty list [];

\For{$W$ in $w$}
{
    $p_{tone}, W$ $\leftarrow$ \FTone{$W$};
    
    $p_{initial}, W$ $\leftarrow$ \FInitial{$W$};
    
    $p_{glide}, W$ $\leftarrow$ \FGlide{$W$};
    
    $p_{vowel}, W$ $\leftarrow$ \FVowel{$W$};
    
    $p_{final}$ $\leftarrow$ \FFinal{$W$};

    phonemes $\leftarrow$ Append $(p_i^{init}, p_i^{glide}, p_i^{vowel}, p_i^{final}, p_i^{tone})$;
}

\Return{phonemes};
\end{algorithm}

\begin{algorithm}[t]
\caption{Algorithm for determining initial of a word.}\label{alg:initial}
\LinesNumbered

\Fn{get\_initial($W$)}{
        \KwIn{A word $W$ in Vietnamese}
        \KwOut{The initial $i$ of $W$}

    $onsets \leftarrow$ [ngh, tr, th, ph, nh, ng, kh, gi, gh, ch, q, đ, x, v, t, s, r, n, m, l, k, h, g, d, c, b];

    $i \leftarrow$ None;

    \For{$onset$ in $onsets$}
    {
        \If{$W$ starts with $onset$}
        {
            \Comment{Words starting with "qu" are kept the onset for later process.}
            \If{$onset \ne k$}
            {
                Remove $onset$ from $W$;
            }

            $i \leftarrow$ IPA character of $onset$ according to Appendix \ref{app:phonetics-orthography};
        }

        \textbf{break};
    }
    \Return{$i, W$};
}
\end{algorithm}

\begin{algorithm}[t]
\caption{Algorithm for determining vowel of a word.}\label{alg:vowel}
\LinesNumbered

\Fn{get\_vowel($W$)}{
        \KwIn{A word $W$ in Vietnamese}
        \KwOut{The vowel $v$ of $W$}

    $nuclei \leftarrow$ [oo, ươ, ưa, uô, ua, iê, yê, ia, ya, e, ê, u, ư, ô, i, y, ơ, â, a, o, ă];

    \For{$nucleus$ in $nuclei$}
    {
        \If{$W$ starts with $nucleus$}
        {
            $v \leftarrow$ IPA character of $nucleus$ according to Appendix \ref{app:phonetics-orthography};

            Remove $nucleus$ from $W$;
            
            \Return{$v$, $W$};
        }
    }

    \Return{None, $W$};
}
\end{algorithm}

We describe in Alg. \ref{alg:text2phoneme} a overview of algorithm of the Base Converter module for converting orthographic form of the transcript to the sequence of phonemes. In this algorithm, we represent each word as a vector of five syllabic components: initial, glide, vowel, final, and tone. From that on, these five phonemes of each word will be mapped to the dialectal phone according to the given province of the audio.

\begin{algorithm}[t]
\caption{Algorithm for determining final of a word.}\label{alg:final}
\LinesNumbered

\Fn{get\_final($W$)}{
        \KwIn{A word $W$ in Vietnamese}
        \KwOut{The final $f$ of $W$}

    $codas \leftarrow$ [ng, nh, ch, u, n, o, p, c, m, y, i, t];

    \If{$W$ in $codas$}
    {
        \Return{$W$};
    }

    \Return{None};
}
\end{algorithm}

\begin{algorithm}[t]
\caption{Algorithm for determining tone of a word.}\label{alg:tone}
\LinesNumbered

\Fn{get\_tone($W$)}{
        \KwIn{A word $W$ in Vietnamese}
        \KwOut{The tone $t$ of $W$}
}
    $t \leftarrow$ tone of word according to Appendix \ref{app:phonetics-orthography};

    Remove tone from $W$;

    \Return{$t, W$};
\end{algorithm}

\begin{algorithm}[t]
\caption{Algorithm for determining glide of a word.}\label{alg:glide}
\LinesNumbered

\Fn{get\_glide($W$)}{
        \KwIn{A word $W$ in Vietnamese}
        \KwOut{The glide $g$ of $W$}

    \If{$W$ starts with \textbf{qu}}
    {
        Remove \textbf{qu} from $W$;
        
        \Return{\textipa{\textsubarch{u}}, $W$};
    }

    \For{$case$ in [oa, oă, oe]}
    {
        \If{$W$ starts with $case$}
        {
            Remove \textbf{o} from $W$;

            \Return{\textipa{\textsubarch{u}}, $W$};
        }
    }

    \For{$case$ in [uê, uy, uơ, ua, uâ, uya]}
    {
        \If{$W$ starts with $case$}
        {
            Remove \textbf{u} from $W$;
            
            \Return{\textipa{\textsubarch{u}}, W}
        }
    }

    \Return{None, W};
}
\end{algorithm}

Moreover, as described in Alg. \ref{alg:tone}, Alg. \ref{alg:initial}, Alg. \ref{alg:glide}, Alg. \ref{alg:vowel}, and Alg. \ref{alg:final}, the computational complexity of the Base Convert (Figure \ref{fig:tokenization}) is linear $\mathcal{O}(n)$ while using a fix sized of vocabulary of phonemes but tokenizing unlimited number of Vietnamese words.

\subsection{Dialectal Converter Module}

Having the phoneme representation for the given transcript, the DiaToken continues to retrieve the dialectal representation of the given transcript according to the name of the province. Vietnam has 63 provinces and their speech sound are grouped into three large dialects as described in Appendix \ref{app:multi-dialect}. DiaToken receives the name of the province then determines the respective dialect and finally convert every phonemes to the corresponding phones. The overall algorithm is described in Alg. \ref{alg:phoneme2phone}.

From Alg. \ref{alg:northern}, Alg. \ref{alg:middle}, and Alg. \ref{alg:southern}, the computational complexity is linear $\mathcal{O}(n)$, which means the complexity of the Dialectal Converter Module is $\mathcal{O}(n)$. Finally, the overall complexity of the DiaToken algorithm is $\mathcal{O}(n)$ which is efficient for tokenizing transcript in the context of multi-dialect.

\SetKwFunction{FNorthern}{get\_northern}
\SetKwFunction{FMiddle}{get\_central}
\SetKwFunction{FSouthern}{get\_southern}

\begin{algorithm}
\caption{The algorithm for converting phonemes to dialectal phones}\label{alg:phoneme2phone}
\KwData{
    \begin{itemize}
        \item The name of the province \textbf{ProID}.
        \item A sequence $p = (p_1, p_2, ..., p_n)$ representing phonemic syllables of the transcript $w = (w_1, w_2, ..., w_n)$. Each $p_i = (p_i^{init}, p_i^{glide}, p_i^{vowel}, p_i^{final}, p_i^{tone})$ is a vector of five syllabic components of word $w_i$.
    \end{itemize}
}
\KwResult{
    A sequence $ph = (ph_1, ph_2, ..., ph_n)$ representing dialectal syllables of the transcript $w = (w_1, w_2, ..., w_n)$. Each $ph_i = (ph_i^{init}, ph_i^{rhyme}, ph_i^{tone})$ is a vector of three dialectal phonetic components of word $w_i$.
}

    $northern \leftarrow$ set of provinces having the Northern dialect according to Appendix \ref{app:multi-dialect};

    $central \leftarrow$ set of provinces having the Central dialect according to Appendix \ref{app:multi-dialect};

    $southern \leftarrow$ set of provinces having the Southern dialect according to Appendix \ref{app:multi-dialect};

    \If{\textbf{ProID} $\in northern$}
    {
        $pth \leftarrow$ \FNorthern($p$);
    }

    \If{\textbf{ProID} $\in central$}
    {
        $pth \leftarrow$ \FMiddle($p$);
    }

    \If{\textbf{ProID} $\in southern$}
    {
        $pth \leftarrow$ \FSouthern($p$);
    }

    \Return{$ph$};
\end{algorithm}

\begin{algorithm*}[htp]
\caption{Algorithm for determining Northern dialect of a syllable.}\label{alg:northern}
\LinesNumbered

\Fn{get\_northern}{
        \KwIn{A sequence of phonemic syllables $s = (s_1, s_2, ..., s_n)$}
        \KwOut{A sequence of Northern phonetic syllables $s = (s_1, s_2, ..., s_n)$}

    \For{$s_i$ in s}
    {
        $s_i^{init} \leftarrow s_i$;

        $s_i^{init} \leftarrow$ IPA character for Northern dialect of the initial $s_i^{init}$ according to Appendix \ref{app:northern-dialect};

        $s_i^{glide} \leftarrow$ IPA character for Northern dialect of the glide $s_i^{glide}$ according to Appendix \ref{app:northern-dialect};

        $s_i^{vowel} \leftarrow$ IPA character for Northern dialect of the vowel $s_i^{vowel}$ according to Appendix \ref{app:northern-dialect};

        $s_i^{final} \leftarrow$ IPA character for Northern dialect of the final $s_i^{final}$ according to Appendix \ref{app:northern-dialect};

        $s_i^{tone}$ $\leftarrow$ IPA character for Northern dialect of the tone $s_i^{tone}$ according to Appendix \ref{app:northern-dialect};

        $s_i^{rhyme} \leftarrow s_i^{glide} \oplus s_i^{vowel} \oplus s_i^{final}$; \Comment{$\oplus$ represents the concatenation operator}

        $s_i \leftarrow (s_i^{init}, s_i^{rhyme}, s_i^{tone})$;
    }
}
\end{algorithm*}

\begin{algorithm*}[htp]
\caption{Algorithm for determining Central dialect of a syllable.}\label{alg:middle}
\LinesNumbered

\Fn{get\_central}{
        \KwIn{A sequence of phonemic syllables $s = (s_1, s_2, ..., s_n)$}
        \KwOut{A sequence of Central phonetic syllables $s = (s_1, s_2, ..., s_n)$}

    \For{$s_i$ in s}
    {
        $s_i^{init} \leftarrow s_i$;

        $s_i^{init} \leftarrow$ IPA character for Central dialect of the initial $s_i^{init}$ according to Appendix \ref{app:central-dialect};

        $s_i^{glide} \leftarrow$ IPA character for Central dialect of the glide $s_i^{glide}$ according to Appendix \ref{app:central-dialect};

        $s_i^{vowel} \leftarrow$ IPA character for Central dialect of the vowel $s_i^{vowel}$ according to Appendix \ref{app:central-dialect};

        $s_i^{final} \leftarrow$ IPA character for Central dialect of the final $s_i^{final}$ according to Appendix \ref{app:central-dialect};

        $s_i^{tone}$ $\leftarrow$ IPA character for Central dialect of the tone $s_i^{tone}$ according to Appendix \ref{app:central-dialect};

        $s_i^{rhyme} \leftarrow s_i^{glide} \oplus s_i^{vowel} \oplus s_i^{final}$; \Comment{$\oplus$ represents the concatenation operator}

        $s_i \leftarrow (s_i^{init}, s_i^{rhyme}, s_i^{tone})$;
    }
}
\end{algorithm*}

\begin{algorithm*}[htp]
\caption{Algorithm for determining Southern dialect of a syllable.}\label{alg:southern}
\LinesNumbered

\Fn{get\_southern}{
        \KwIn{A sequence of phonemic syllables $s = (s_1, s_2, ..., s_n)$}
        \KwOut{A sequence of Southern phonetic syllables $s = (s_1, s_2, ..., s_n)$}

    \For{$s_i$ in s}
    {
        $s_i^{init} \leftarrow s_i$;

        $s_i^{init} \leftarrow$ IPA character for Southern dialect of the initial $s_i^{init}$ according to Appendix \ref{app:southern-dialect};

        $s_i^{glide} \leftarrow$ IPA character for Southern dialect of the glide $s_i^{glide}$ according to Appendix \ref{app:southern-dialect};

        $s_i^{vowel} \leftarrow$ IPA character for Southern dialect of the vowel $s_i^{vowel}$ according to Appendix \ref{app:southern-dialect};

        $s_i^{final} \leftarrow$ IPA character for Southern dialect of the final $s_i^{final}$ according to Appendix \ref{app:southern-dialect};

        $s_i^{tone}$ $\leftarrow$ IPA character for Southern dialect of the tone $s_i^{tone}$ according to Appendix \ref{app:southern-dialect};

        $s_i^{rhyme} \leftarrow s_i^{glide} \oplus s_i^{vowel} \oplus s_i^{final}$; \Comment{$\oplus$ represents the concatenation operator}

        $s_i \leftarrow (s_i^{init}, s_i^{rhyme}, s_i^{tone})$;
    }
}
\end{algorithm*}

\end{document}